%% file: atari.tex
\documentclass[twoside,11pt]{article}
\usepackage{jair,rawfonts}
\usepackage{theapa}

\usepackage{amsmath}
\usepackage{algorithm}
\usepackage{algorithmic}

\usepackage{graphicx}
\usepackage{subfigure}
\usepackage{url}

\usepackage{booktabs}

\newif\ifjair
\jairtrue

\ifjair
\jairheading{47}{2013}{253--279}{02/13}{06/13}
\fi

\ShortHeadings{The Arcade Learning Environment: An Evaluation Platform for General Agents}
{Bellemare, Naddaf, Veness, \& Bowling}
\firstpageno{253}

\newcommand{\gamename}[1]{{\sc #1}}
\usepackage{amssymb}
\newcommand{\bR}{\mathbb{R}}

\newcommand{\indic}[1]{\mathbb{I}_{[#1]}}

\begin{document}

\title{The Arcade Learning Environment:\\An Evaluation Platform for General Agents}

\author{\name Marc G. Bellemare \email mg17@cs.ualberta.ca \\
\addr University of Alberta, Edmonton, Alberta, Canada 
\AND
\name Yavar Naddaf \email yavar@empiricalresults.ca \\
\addr Empirical Results Inc., Vancouver, \\
British Columbia, Canada 
\AND
\name Joel Veness \email veness@cs.ualberta.ca \\
\name Michael Bowling \email bowling@cs.ualberta.ca \\
\addr University of Alberta, Edmonton, Alberta, Canada}

% For research notes, remove the comment character in the line below.
% \researchnote

\ifjair\else
\thispagestyle{empty}
\fi

\maketitle

\begin{abstract}
In this article we introduce the Arcade Learning Environment (ALE): both a challenge problem and a platform and methodology for evaluating the development of general, domain-independent AI technology.  ALE provides an interface to hundreds of Atari 2600 game environments, each one different, interesting, and designed to be a challenge for human players.  ALE presents significant research challenges for reinforcement learning, model learning, model-based planning, imitation learning, transfer learning, and intrinsic motivation.  Most importantly, it provides a rigorous testbed for evaluating and comparing approaches to these problems.  We illustrate the promise of ALE by developing and benchmarking domain-independent agents designed using well-established AI techniques for both reinforcement learning and planning.  In doing so, we also propose an evaluation methodology made possible by ALE, reporting empirical results on over 55 different games.  All of the software, including the benchmark agents, is publicly available.
\end{abstract}

\section{Introduction}

A longstanding goal of artificial intelligence is the development of algorithms capable of general competency in a variety of tasks and domains without the need for domain-specific tailoring. To this end, different theoretical frameworks have been proposed to formalize the notion of ``big'' artificial intelligence \cite<e.g.,>{Russell97rationalityand,Hutter:04uaibook,legg08machine}.
Similar ideas have been developed around the theme of \emph{lifelong learning}: learning a reusable, high-level understanding of the world from raw sensory data \shortcite{thrun95lifelong,pierce_kuipers_97,stober08pixels,sutton11horde}. The growing interest in competitions such as the General Game Playing competition \cite{ggpcomp}, Reinforcement Learning competition \cite{whiteson10}, and the International Planning competition \shortcite{coles_12} also suggests the artificial intelligence community's desire for the emergence of algorithms that provide general competency.

Designing generally competent agents raises the question of how to best evaluate them.
Empirically evaluating general competency on a handful of parametrized benchmark problems is, by definition, flawed. Such an evaluation is prone to method overfitting \cite{whiteson11} and discounts the amount of expert effort necessary to transfer the algorithm to new domains. Ideally, the algorithm should be compared across domains that are (i) {\em varied} enough to claim generality, (ii) each {\em interesting} enough to be representative of settings that might be faced in practice, and (iii) each created by an {\em independent} party to be free of experimenter's bias.

In this article, we introduce the Arcade Learning Environment (ALE): a new challenge problem, platform, and experimental methodology for empirically assessing agents designed for general competency.
ALE is a software framework for interfacing with emulated Atari 2600 game environments.  The Atari 2600, a second generation game console, was originally released in 1977 and remained massively popular for over a decade. 
Over 500 games were developed for the Atari 2600, spanning a diverse range of genres such as shooters, beat'em ups, puzzle, sports, and action-adventure games; many game genres were pioneered on the console. 
While modern game consoles involve visuals, controls, and a general complexity that rivals the real world, Atari 2600 games are far simpler.
In spite of this, they still pose a variety of challenging and interesting situations for human players.

ALE is both an experimental methodology and a challenge problem for general AI competency. 
In machine learning, it is considered poor experimental practice to both train and evaluate an algorithm on the same data set, as it can grossly over-estimate the algorithm's performance.  The typical practice is instead to train on a {\em training set} then evaluate on a disjoint {\em test set}.
With the large number of available games in ALE, we propose that a similar methodology can be used to the same effect: an approach's domain representation and parametrization should be first tuned on a small number of {\em training games}, before testing the approach on unseen {\em testing games}. Ideally, agents designed in this fashion are evaluated on the testing games only once, with no possibility for subsequent modifications to the algorithm. 
While general competency remains the long-term goal for artificial intelligence, ALE proposes an achievable stepping stone: techniques for general competency across the gamut of Atari 2600 games.  We believe this represents a goal that is attainable in a short time-frame yet formidable enough to require new technological breakthroughs.

\section{Arcade Learning Environment}
\label{sec:atari}

We begin by describing our main contribution, the Arcade Learning Environment (ALE).  ALE is a software framework designed to make it easy to develop agents that play arbitrary Atari 2600 games.

\subsection{The Atari 2600} 

The Atari 2600 is a home video game console developed in 1977 and sold for over a decade \cite{MontfortBogost2009}.  It popularized the use of general purpose CPUs in game console hardware, with game code distributed through cartridges.  Over 500 original games were released for the console; ``homebrew'' games continue to be developed today, over thirty years later. The console's joystick, as well as some of the original games such as \gamename{Adventure} and \gamename{Pitfall!}, are iconic symbols of early video games.  Nearly all arcade games of the time -- \gamename{Pac-Man} and \gamename{Space Invaders} are two well-known examples -- were ported to the console.

\begin{figure}[t!]
\begin{center}
\includegraphics[width=2.5in]{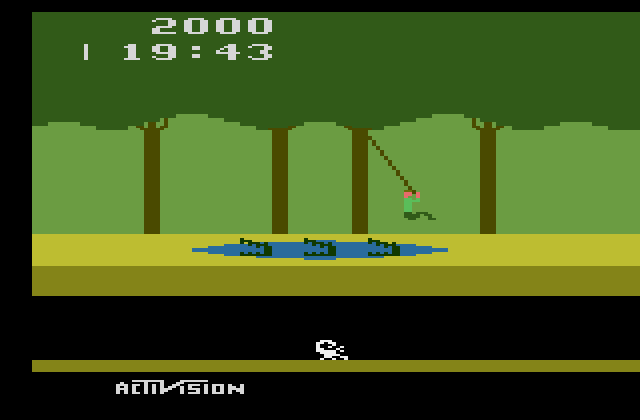}
\includegraphics[width=2.5in]{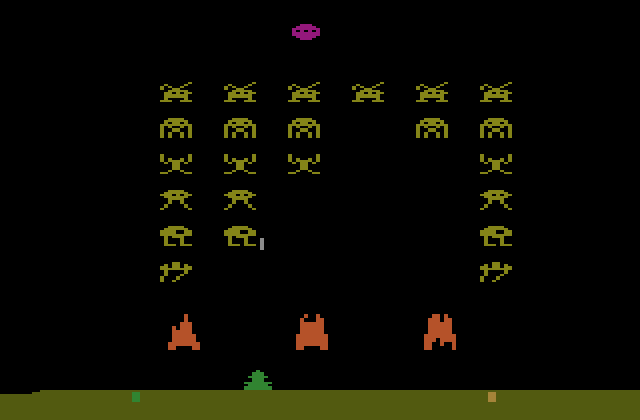}
\caption{Screenshots of \gamename{Pitfall!} and \gamename{Space Invaders}.\label{fig:atari_domain:screenshots}}
\end{center}
\vspace{-1.5em}
\end{figure}

Despite the number and variety of games developed for the Atari 2600, the hardware is relatively simple.  It has a 1.19Mhz CPU and can be emulated much faster than real-time on modern hardware.  The cartridge ROM (typically 2--4kB) holds the game code, while the console RAM itself only holds 128 bytes (1024 bits). A single game screen is 160 pixels wide and 210 pixels high, with a 128-colour palette; 18 ``actions'' can be input to the game via a digital joystick: three positions of the joystick for each axis, plus a single button.  The Atari 2600 hardware limits the possible complexity of games, which we believe strikes the perfect balance: a challenging platform offering conceivable near-term advancements in learning, modelling, and planning.

\subsection{Interface}
ALE is built on top of Stella\footnote{\url{http://stella.sourceforge.net/}}, an open-source Atari 2600 emulator. 
It allows the user to interface with the Atari 2600 by receiving joystick motions, sending screen and/or RAM information, and emulating the platform. 
ALE also provides a game-handling layer which transforms each game into a standard reinforcement learning problem by identifying the accumulated score and whether the game has ended. By default, each observation consists of a single game screen (frame): a 2D array of 7-bit pixels, 160 pixels wide by 210 pixels high. 
The action space consists of the 18 discrete actions defined by the joystick controller.  The game-handling layer also specifies the minimal set of actions needed to play a particular game, although none of the results in this paper make use of this information.
When running in real-time, the simulator generates 60 frames per second, and at full speed emulates up to 6000 frames per second.
The reward at each time-step is defined on a game by game basis, typically by taking the difference in score or points between frames. 
An episode begins on the first frame after a reset command is issued, and terminates when the game ends. 
The game-handling layer also offers the ability to end the episode after a predefined number of frames\footnote{This functionality is needed for a small number of games to ensure that they always terminate.
This prevents situations such as in \gamename{Tennis}, where a degenerate agent could choose to play indefinitely by refusing to serve.}.
The user therefore has access to several dozen games through a single common interface, and adding 
support for new games is relatively straightforward.

ALE further provides the functionality to save and restore the state of the emulator. 
When issued a \emph{save-state} command, ALE saves all the relevant data about the current game, including the contents of the RAM, registers, and address counters. 
The \emph{restore-state} command similarly resets the game to a previously saved state. 
This allows the use of ALE as a generative model to study topics such as planning and model-based reinforcement learning.

\subsection{Source Code}
ALE is released as free, open-source software under the terms of the GNU General Public License.
The latest version of the source code is publicly available at:
\begin{quote} 
\url{http://arcadelearningenvironment.org}
\end{quote}
The source code for the agents used in the benchmark experiments below is also available on the publication page for this article on the same website. While ALE itself is written in C++, a variety of interfaces are available that allow users to interact with ALE in the programming language of their choice. Support for new games is easily added by implementing a derived class representing the game's particular reward and termination functions. 

\section{Benchmark Results}
Planning and reinforcement learning are two different AI problem formulations that can naturally be investigated within the ALE framework. Our purpose in presenting benchmark results for both of these formulations is two-fold.  First, these results provide a baseline performance for traditional techniques, establishing a point of comparison with future, more advanced, approaches.  Second, in describing these results we illustrate our proposed methodology for doing empirical validation with ALE.

\subsection{Reinforcement Learning}
\label{sec:RL}

We begin by providing benchmark results using SARSA$(\lambda)$, a traditional technique for model-free reinforcement learning.
Note that in the reinforcement learning setting, the agent does not have access to a model of the game dynamics.
At each time step, the agent selects an action and receives a reward and an observation, and the agent's aim is to maximize its accumulated reward. 
In these experiments, we augmented the SARSA($\lambda$) algorithm with linear function approximation, replacing traces, and $\epsilon$-greedy exploration. 
A detailed explanation of SARSA($\lambda$) and its extensions can be found in the work of \citeA{sutton_barto_98}. 

\subsubsection{Feature Construction}
In our approach to the reinforcement learning setting, the most important design issue is the choice of features to use with linear function approximation.
We ran experiments using five different sets of features, which we now briefly explain;
a complete description of these feature sets is given in Appendix~\ref{apdx:feature_sets}. Of these
sets of features, BASS, DISCO and RAM were originally introduced by \citeA{naddaf2010}, while the rest are novel.

\paragraph{Basic.}
The \emph{Basic} method, derived from \citeauthor{naddaf2010}'s BASS \citeyear{naddaf2010}, encodes the presence of colours on the Atari 2600 screen.
The Basic method first removes the image background by storing the frequency of colours at each pixel location within a histogram. 
Each game background is precomputed offline, using 18,000 observations collected from sample trajectories. The sample trajectories are generated by following a human-provided trajectory for a random number of steps and subsequently selecting actions uniformly at random.  
The screen is then divided into $16 \times 14$ tiles. 
Basic generates one binary feature for each of the $128$ colours and each of the tiles, giving a total of 28,672 features.

\paragraph{BASS.} The \emph{BASS} method behaves identically to the Basic method save in two respects. First, BASS augments the Basic feature set with pairwise combinations of its features. Second, BASS uses a smaller, 8-colour encoding to ensure that the number of pairwise combinations remains tractable. 

\paragraph{DISCO.} The \emph{DISCO} method aims to detect objects within the Atari 2600 screen. To do so, it first preprocesses 36,000 observations from sample trajectories generated as in the \emph{Basic} method. DISCO also performs the background subtraction steps as in Basic and BASS. Extracted objects are then labelled into classes. During the actual training, DISCO infers the class label of detected objects and encodes their position and velocity using tile coding \cite{sutton_barto_98}.

\paragraph{LSH.} The \emph{LSH} method maps raw Atari 2600 screens into a small set of binary features
using Locally Sensitive Hashing \cite{gionis_99}. The screens are mapped using random 
projections, such that visually similar screens are more likely to generate the same features.

\paragraph{RAM.} The \emph{RAM} method works on an entirely different observation space than the other four methods.  Rather than receiving in Atari 2600 screen as an observation, it directly observes the Atari 2600's 1024 bits of memory.  
Each bit of RAM is provided as a binary feature together with the pairwise logical-AND of every pair of bits.

\subsubsection{Evaluation Methodology}
\label{sec:RL:experimental_setup}

We first constructed two sets of games, one for training and the other for testing. 
We used the training games for parameter tuning as well as design refinements, and the testing games for the final evaluation of our methods.
Our training set consisted of five games: \gamename{Asterix}, \gamename{Beam Rider}, \gamename{Freeway}, \gamename{Seaquest} and \gamename{Space Invaders}. 
The parameter search involved finding suitable values for the parameters to the SARSA($\lambda$) algorithm, i.e. the learning rate, exploration rate, discount factor, and the decay rate $\lambda$. 
We also searched the space of feature generation parameters, for example the abstraction level for the BASS agent. 
The results of our parameter search are summarized in Appendix~\ref{appendix:parameters}.
Our testing set was constructed by choosing semi-randomly from the 381 games listed on Wikipedia\footnote{http://en.wikipedia.org/wiki/List\_of\_Atari\_2600\_games (July 12, 2012)} at the time of writing. 
Of these games, 123 games have their own Wikipedia page, have a single player mode, are not adult-themed or prototypes, and can be emulated in ALE. From this list,
50 games were chosen at random to form the test set.

Evaluation of each method on each game was performed as follows.
An \emph{episode} starts on the frame that follows the reset command, and terminates when the end-of-game condition is detected or after 5 minutes of real-time play (18,000 frames), whichever comes first. 
During an episode, the agent acts every 5 frames, or equivalently 12 times per second of gameplay. 
A reinforcement learning \emph{trial} consists of 5,000 training episodes, followed by 500 evaluation episodes during which no learning takes place. 
The agent's performance is measured as the average score achieved during the evaluation episodes. 
For each game, we report our methods' average performance across 30 trials. 

For purposes of comparison, we also provide performance measures for three simple baseline agents -- \emph{Random}, \emph{Const} and \emph{Perturb} -- as well as the performance of a non-expert human player. The Random agent picks a random action on every frame. The Const agent selects a single fixed action throughout an episode; our results reflect the highest score achieved by any single action within each game. The Perturb agent selects a fixed action with probability 0.95 and otherwise acts uniformly randomly; for each game, we report the performance of the best policy of this type. 
Additionally, we provide human player results that report the five-episode average score obtained by a beginner (who had never previously played Atari 2600 games) playing selected games. Our aim is not to provide exhaustive or accurate human-level benchmarks, which would be beyond the scope of this paper, but rather to offer insight into the performance level achieved by our agents. 

\subsubsection{Results}
\label{sec:RL:results}

\begin{table}
\small
\begin{center}
\include{tables/selectedRLResults}
\end{center}
\vspace{-1.6em}
\caption{Reinforcement Learning results for selected games. \gamename{Asterix} and \gamename{Seaquest} are part of the training set.\label{table:selected_rl_results}}
\end{table}

A complete report of our reinforcement learning results is given in Appendix 
\ref{appendix:detailed_results}. Table \ref{table:selected_rl_results} shows a small subset of 
results from two training games and three test games.  In 40 games out of 55, learning agents perform 
better than the baseline agents.  In some games, e.g., \gamename{Double Dunk}, \gamename{Journey Escape} and \gamename{Tennis}, the no-action baseline policy performs the best by essentially refusing to play and thus incurring no negative reward.  Within the 40 games for which learning occurs, the BASS method
generally performs best. DISCO performed particularly poorly compared to the other learning methods. 
The RAM-based agent, surprisingly, did not outperform image-based methods, despite building its representation from raw game {\em state}. It appears the screen image carries structural information that is not easily extracted from the RAM bits.

Our reinforcement learning results show that while some learning progress is already possible in Atari 2600 games, much more work remains to be done. 
Different methods perform well on different games, and no single method performs well on all games. 
Some games are particularly challenging.  For example, platformers such as \gamename{Montezuma's Revenge} seem to require high-level planning far beyond what our current, domain-independent methods provide.  \gamename{Tennis} requires fairly elaborate behaviour before observing any positive reward, but simple behaviour can avoid negative rewards.  Our results also highlight the value of ALE as an experimental methodology.
For example, the DISCO approach performs reasonably well on the training set, but suffers a dramatic reduction in performance when applied to unseen games.
This suggests the method is less robust than the other methods we studied.  After a quick glance at the full table of results in Appendix~\ref{appendix:detailed_results}, it is clear that summarizing results across such varied domains needs further attention; we explore this issue further in Section \ref{sec:comparison}. 

\subsection{Planning}
\label{sec:planning}

The Arcade Learning Environment can naturally be used to study planning techniques by using the emulator itself as a generative model. 
Initially it may seem that allowing the agent to plan into the future with a perfect model trivializes the problem.
However, this is not the case: the size of state space in Atari 2600 games prohibits exhaustive search. 
Eighteen different actions are available at every frame; at 60 frames per second, looking ahead one second requires $18^{60} \approx 10^{75}$ simulation steps.
Furthermore, rewards are often sparsely distributed, which causes significant horizon effects in many search algorithms. 

\subsubsection{Search Methods}
\label{sec:planning:methods}

We now provide benchmark ALE results for two traditional search methods.
Each method was applied online to select an action at every time step (every five frames) until the game was over.

\paragraph{Breadth-first Search.}
\label{sec:agents:search:fulltree}
Our first approach builds a search tree in a breadth-first fashion until a node limit is reached.
Once the tree is expanded, node values are updated recursively from the bottom of the tree to the root. The agent then selects the action corresponding to the branch with the highest discounted sum of rewards. Expanding the full search tree requires a large number of simulation steps. 
For instance, selecting an action every 5 frames and allowing a maximum of 100,000 simulation steps per frame, the agent can only look ahead about a third of a second. 
In many games, this allows the agent to collect immediate rewards and avoid death but little else. For example, in \gamename{Seaquest} the agent must collect a swimmer and return to the surface before running out of air, which involves planning far beyond one second.

\paragraph{UCT: Upper Confidence Bounds Applied to Trees.}
\label{sec:agents:search:uct}
A preferable alternative to exhaustively expanding the tree is to simulate deeper into the more promising branches. To do this, we need to find a balance between expanding the higher-valued branches and spending simulation steps on the lower-valued branches to get a better estimate of their values. 
The UCT algorithm, developed by \citeA{kocsis_06}, deals with the exploration-exploitation dilemma by treating each node of a search tree as a multi-armed bandit problem. UCT uses a variation of UCB1, a bandit algorithm, to choose which child node to visit next. A common practice is to apply a $t$-step random simulation at the end of each leaf node to obtain an estimate from a longer trajectory. By expanding the more valuable branches of the tree and carrying out a random simulation at the leaf nodes, UCT is known to perform well in many different settings \shortcite{mcts_survery2012}.
 
Our UCT implementation was entirely standard, except for one optimization.
Few Atari games actually distinguish between all 18 actions at every time step. 
In \gamename{Beam Rider}, for example, the down action does nothing, and pressing the button when a bullet has already been shot has no effect. We exploit this fact as follows: after expanding the children of a node in the search tree, we compare the resulting emulator states. Actions that result in the same state are treated as duplicates and only one of the actions is considered in the search tree. This reduces the branching factor, thus allowing deeper search. At every step, we also reuse the part of our search tree corresponding to the selected action. Pseudocode for our implementation of the UCT algorithm is given in Appendix \ref{appendix:uct_pseudocode}.

\subsubsection{Experimental Setup}
\label{sec:exp}

We designed and tuned our algorithms based on the same five training games used in Section \ref{sec:RL}, and subsequently evaluated the methods on the fifty games of the testing set. 
The training games were used to determine the length of the search horizon as well as the constant controlling the amount of exploration at internal nodes of the tree. 
Each episode was set to last up to 5 minutes of real-time play (18,000 frames), with actions selected every 5 frames, matching our settings in Section \ref{sec:RL:experimental_setup}. On average, each action selection step took on the order of 15 seconds. 
We also used the same discount factor as in Section \ref{sec:RL}.
We ran our algorithms for 10 episodes per game.  Details of the algorithmic parameters can be found in Appendix \ref{appendix:parameters}.

\subsubsection{Results}\label{sec:planning:results}

\begin{table}
\small
\begin{center}
\include{tables/selectedSearchResults}
\end{center}
\vspace{-1.6em}
\caption{Results for selected games. \gamename{Asterix} and \gamename{Seaquest} are part of the training set.\label{table:selected_search_results}}
\end{table}

A complete report of our search results is given in Appendix \ref{appendix:detailed_results}.
Table \ref{table:selected_search_results} shows results on a selected subset of games.
For reference purposes, we also include the performance of the best learning agent and the best 
baseline policy from Table \ref{table:selected_rl_results}.
Together, our two search methods performed better than both learning agents and the baseline policies on 49 of 55 games. In most cases, UCT performs significantly better than breadth-first search.
Four of the six games for which search methods do not perform best are games where rewards are sparse and require long-term planning. These are \gamename{Freeway}, \gamename{Private Eye}, \gamename{Montezuma's Revenge} and \gamename{Venture}. 

\section{Evaluation Metrics for General Atari 2600 Agents}
\label{sec:comparison}

Applying algorithms to a large set of games as we did in Sections \ref{sec:RL} and 
\ref{sec:planning} presents difficulties when interpreting the results.
While the agent's goal in all games is to maximize its score, scores for two different games cannot be easily compared. 
Each game uses its own scale for scores, and different game mechanics make some games harder to learn than others. 
The challenges associated with comparing general agents has been previously highlighted by \citeA{whiteson11}.
Although we can always report full performance tables, as we did in Appendix \ref{appendix:detailed_results}, some more compact summary statistics are also desirable.
We now introduce some simple metrics that help compare agents across a diverse set of domains, such as our test set of Atari 2600 games.

\subsection{Normalized Scores}

Consider the scores $s_{g,1}$ and $s_{g,2}$ achieved by two algorithms in game $g$. Our goal here is to explore methods that allow us to compare two sets of scores $S_1 = \{ s_{g_1, 1}, \dots, s_{g_n, 1} \}$ and $S_2 = \{ s_{g_1, 2}, \dots, s_{g_n, 2} \}$. The approach we take is to transform $s_{g,i}$ into a normalized score $z_{g,i}$ with the aim of comparing normalized scores across games; in the ideal case, $z_{g,i} = z_{g',i}$ implies that algorithm $i$ performs as well on game $g$ as on game $g'$. In order to compare algorithms over a set of games, we aggregate normalized scores for each game and each algorithm. 

The most natural way to compare games with different scoring scales is to normalize scores so that the numerical values become comparable. All of our normalization methods are defined using the notion of a  \emph{score range} $[r_{g,\min}, r_{g,\max} ]$ computed for each game. Given such a score range, score $s_{g,i}$ is normalized by computing $z_{g,i} := (s_{g,i} - r_{g,\min}) \; / \; (r_{g,\max} - r_{g,\min})$.

\begin{figure}
\begin{center}
\includegraphics[width=2in]{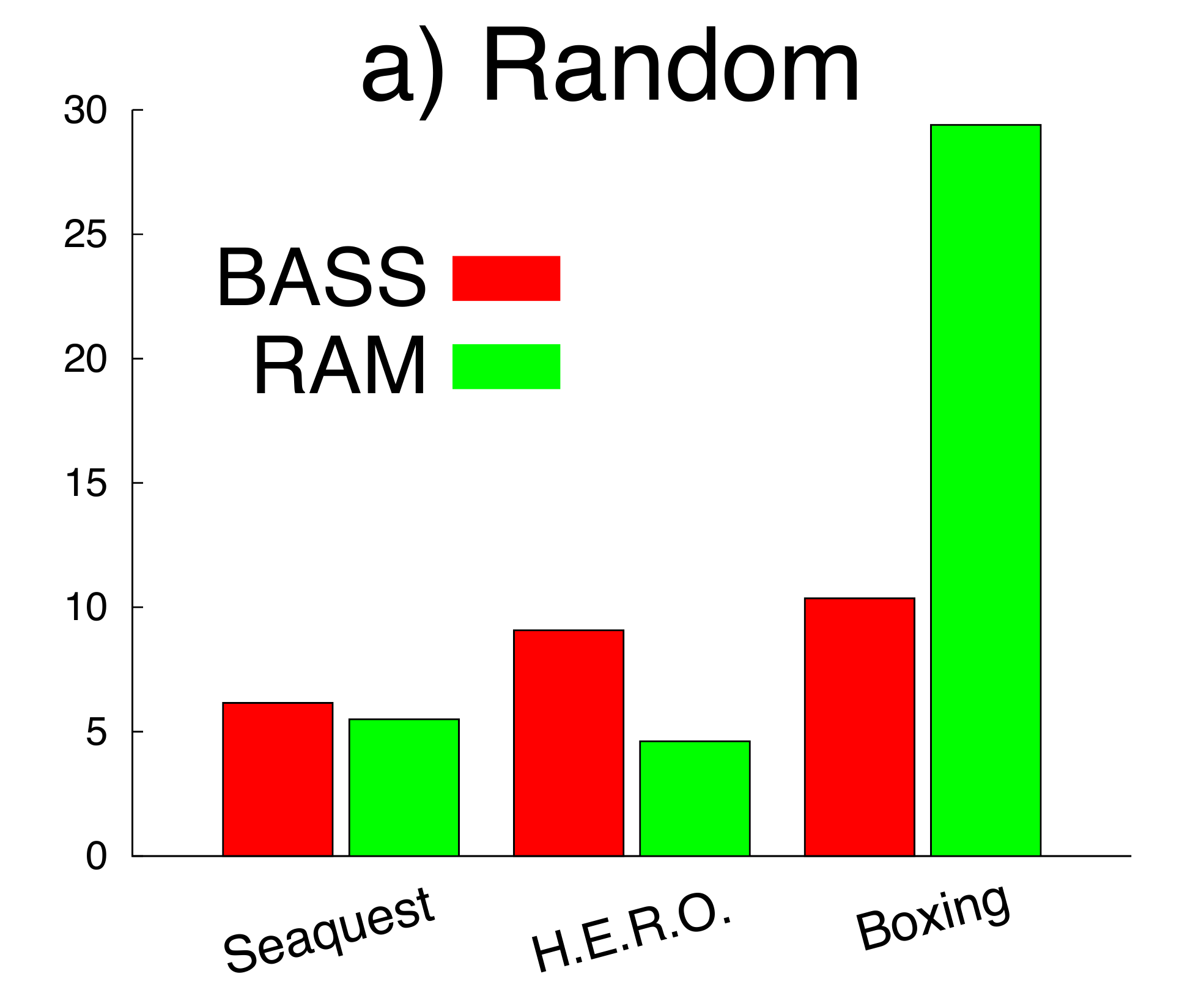}
\hspace{-1em}
\includegraphics[width=2in]{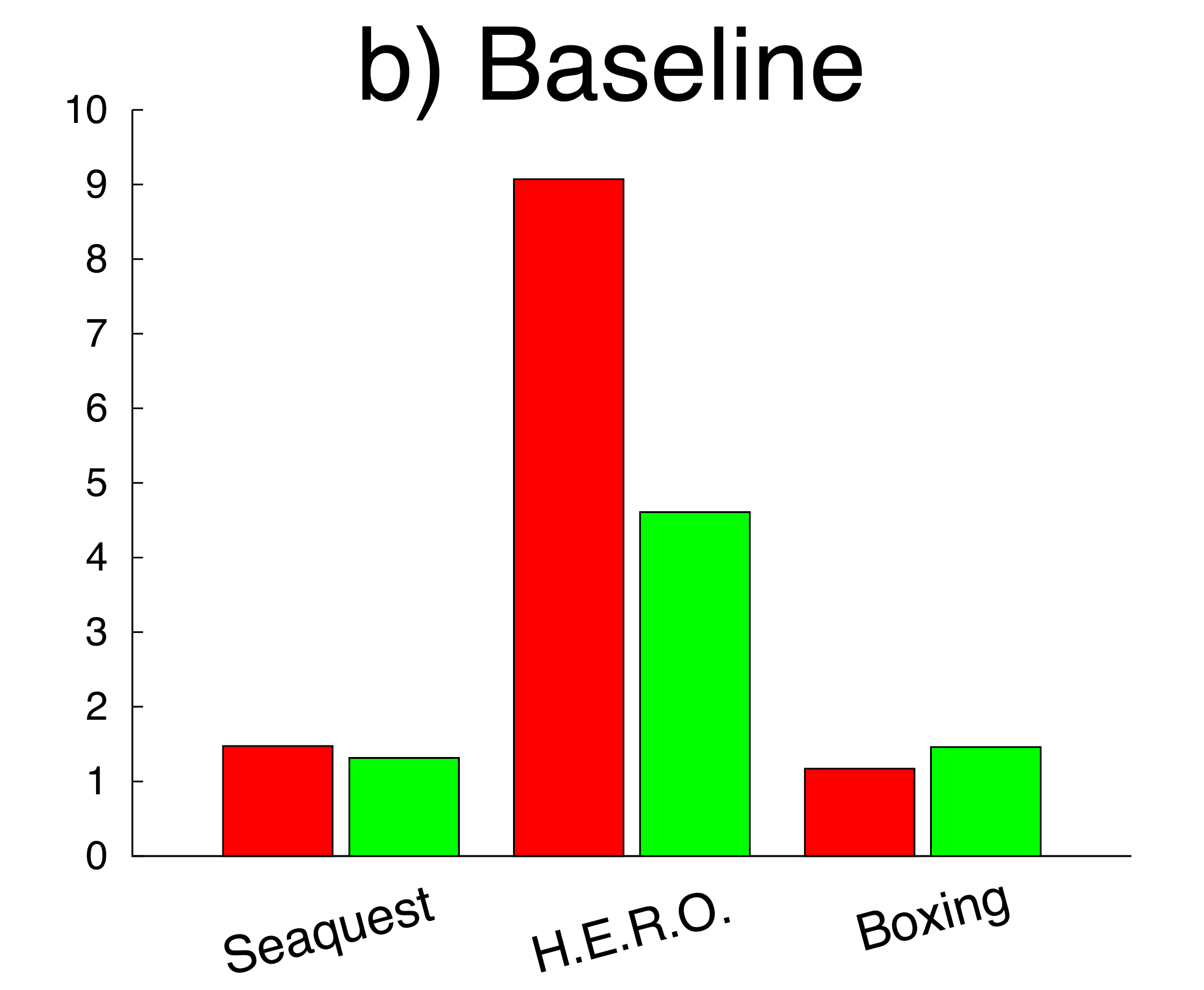}
\hspace{-1em}
\includegraphics[width=2in]{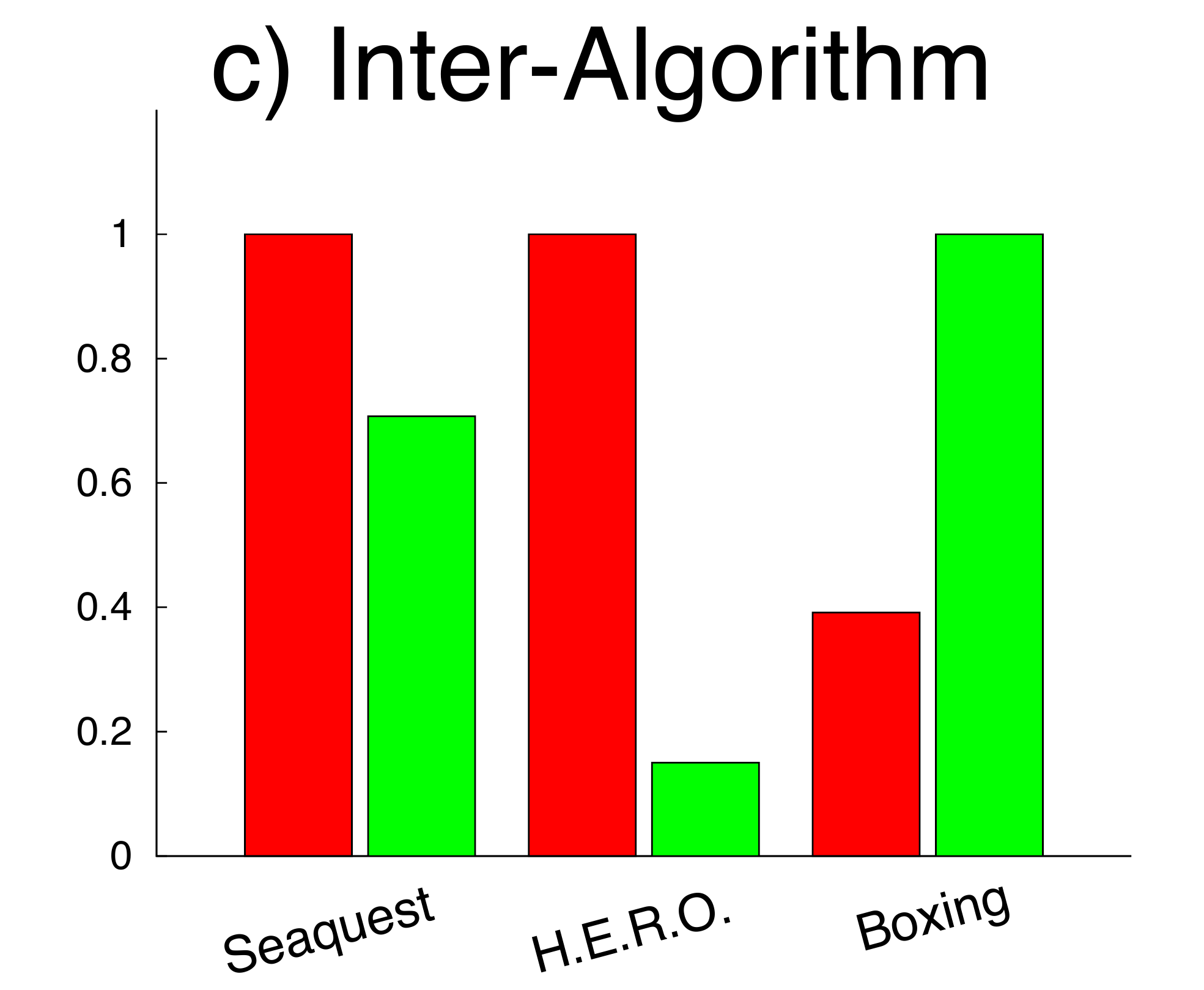}
\caption{Left to right: random-normalized, baseline and inter-algorithm scores.\label{fig:evaluation:normalized_scores}}
\end{center}
\end{figure}

\subsubsection{Normalization to a Reference Score}

One straightforward method is to normalize to a score range defined by repeated runs of a random agent across each game. Here, $r_{g,\max}$ is the absolute value of the average score achieved by the random agent, and $r_{g,\min} = 0$. 
Figure \ref{fig:evaluation:normalized_scores}a depicts the \emph{random-normalized scores} achieved by BASS and RAM on three games.  Two issues arise with this approach: the scale of normalized scores may be excessively large and normalized scores are generally not translation invariant. 
The issue of scale is best seen in a game such as \gamename{Freeway}, for which the random agent achieves a score close to 0: scores achieved by learning agents, in the 10-20 range, are normalized into thousands. 
By contrast, no learning agent achieves a random-normalized score greater than 1 in Asteroids. 

\subsubsection{Normalizing to a Baseline Set}

Rather than normalizing to a single reference we may normalize to the score range implied by
a set of references. Let $b_{g,1}, \dots, b_{g,k}$ be a set of reference scores. A 
method's \emph{baseline score} is computed using the score range 
$[ \min_{i \in \{ 1, \dots, k \} } b_{g,i}, \max_{i \in \{1, \dots, k\}} b_{g,i} ]$.

%$}}$
% The above is here because mvim is getting confused about brackets 

Given a sufficiently rich set of reference scores, baseline normalization allows us to reduce the scores for most games to comparable quantities, and lets us know whether meaningful performance was obtained. 
Figure \ref{fig:evaluation:normalized_scores}b shows example baseline scores. The score range for these scores corresponds to the scores achieved by 37 baseline agents (Section \ref{sec:RL:experimental_setup}): \emph{Random}, \emph{Const} (one policy per action), and \emph{Perturb} (one policy per action). 

A natural idea is to also include scores achieved by human players into the baseline set. 
For example, one may include the score achieved by an expert as well as the score achieved by a beginner. However, using human scores raises its own set of issues. 
For example, humans often play games without seeking to maximize score; humans also benefit from prior knowledge that is difficult to incorporate into domain-independent agents.

\subsubsection{Inter-Algorithm Normalization}

A third alternative is to normalize using the scores achieved by the algorithms themselves. 
Given $n$ algorithms, each achieving score $s_{g,i}$ on game $g$, we define the 
\emph{inter-algorithm score} using the score range 
$[ \min_{i \in \{ 1, \dots, n \} } s_{g,i}, \max_{i \in \{1, \dots, n\}} s_{g,i} ]$.
%$}}$
% The above is here because mvim is getting confused about brackets 
By definition, $z_{g,i} \in [0, 1]$. A special case of this is when n=2, where 
$z_{g,i} \in \{0, 1\}$ indicates which algorithm is better than the other.
Figure \ref{fig:evaluation:normalized_scores}c shows example inter-algorithm scores; the 
relevant score ranges are constructed from the performance of all five learning agents. 

Because inter-algorithm scores are bounded, this type of normalization is an appealing solution to compare the relative performance of different methods.
Its main drawback is that it gives no indication of the objective performance of the best algorithm. 
A good example of this is \gamename{Venture}: the inter-algorithm score of 1.0 achieved by BASS does not reflect the fact that none of our agents achieved a score remotely comparable to a human's performance. The lack of objective reference in inter-algorithm normalization suggests that it should be used to complement other scoring metrics. 

\subsection{Aggregating Scores}

Once normalized scores are obtained for each game, the next step is to produce a measure that reflects how well each agent performs across the set of games. As illustrated by Table \ref{table:appendix:detailed_results:rl}, a large table of numbers does not easily permit comparison between algorithms. We now describe three methods to aggregate normalized scores.

\begin{figure}
\begin{center}
\includegraphics[width=2in]{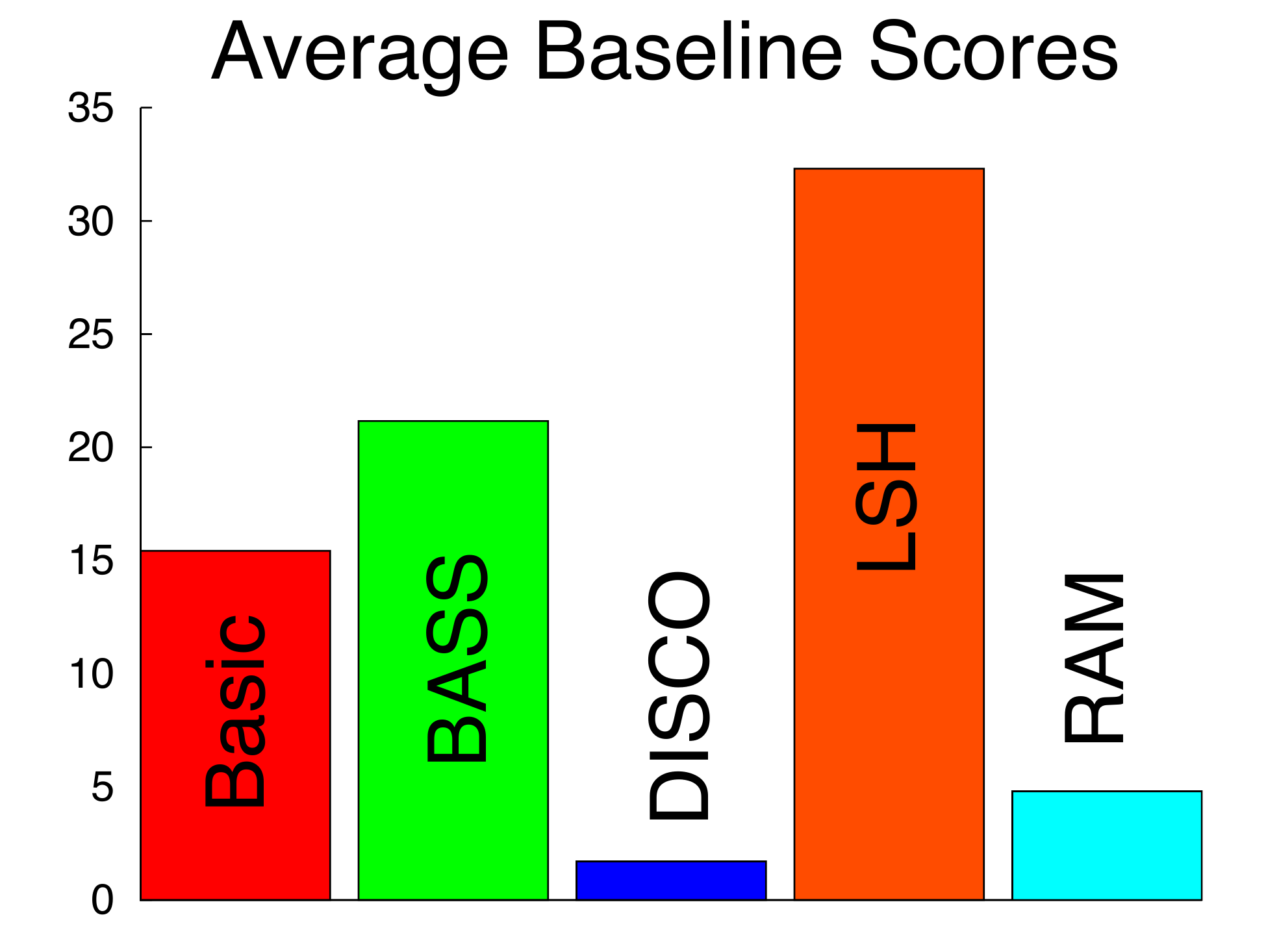}
\includegraphics[width=2in]{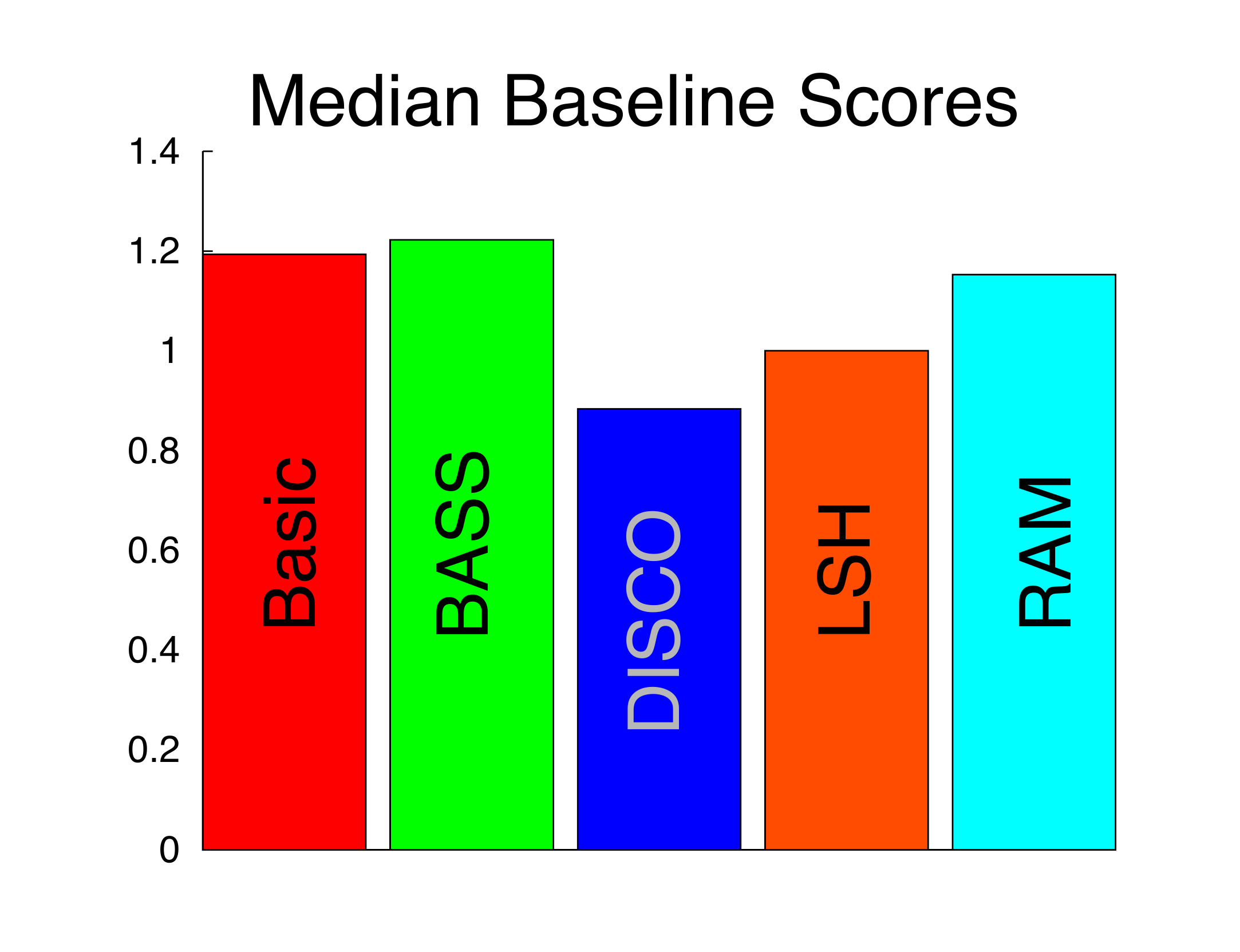}
\includegraphics[width=2in]{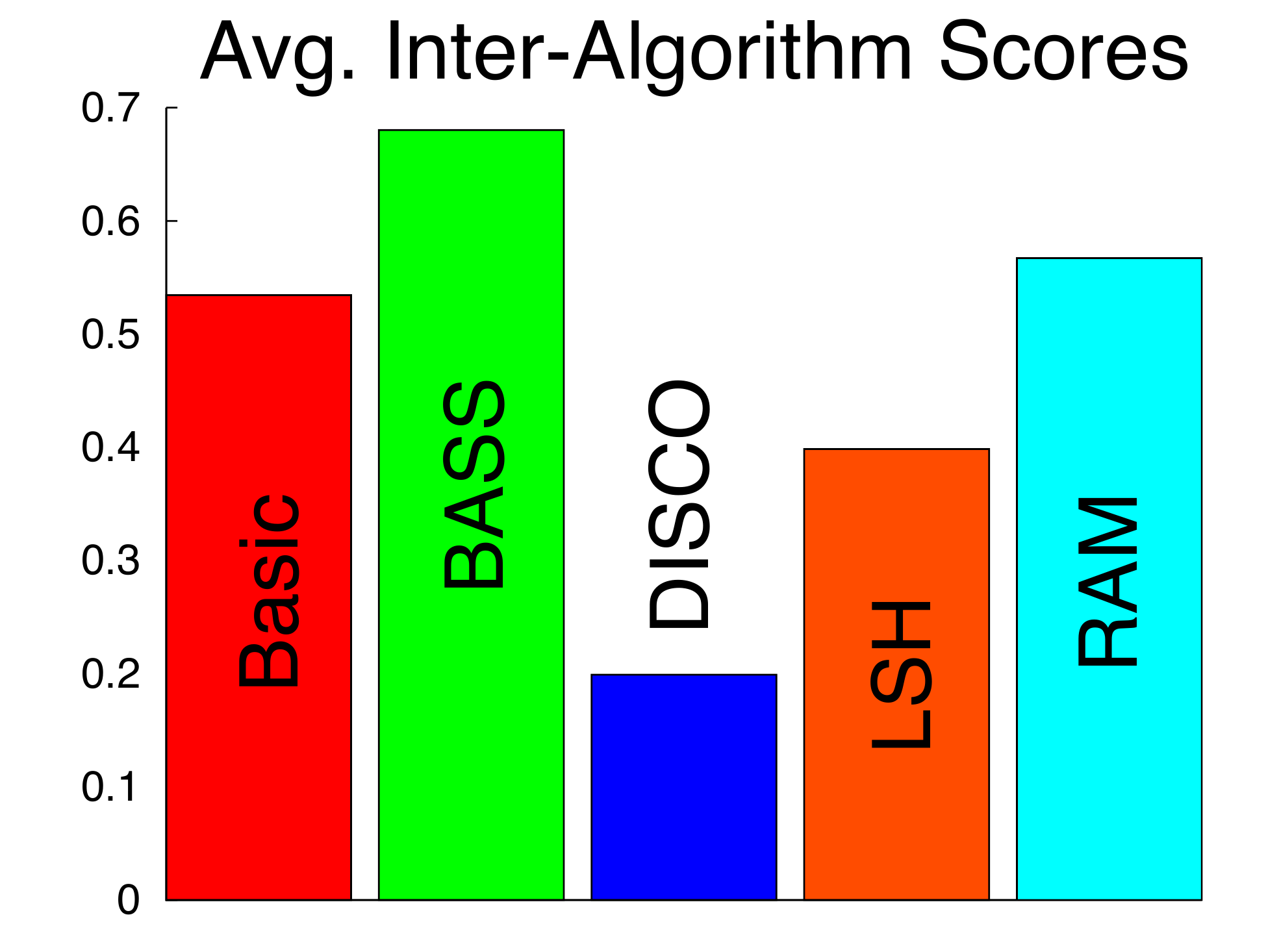}
\includegraphics[width=2in]{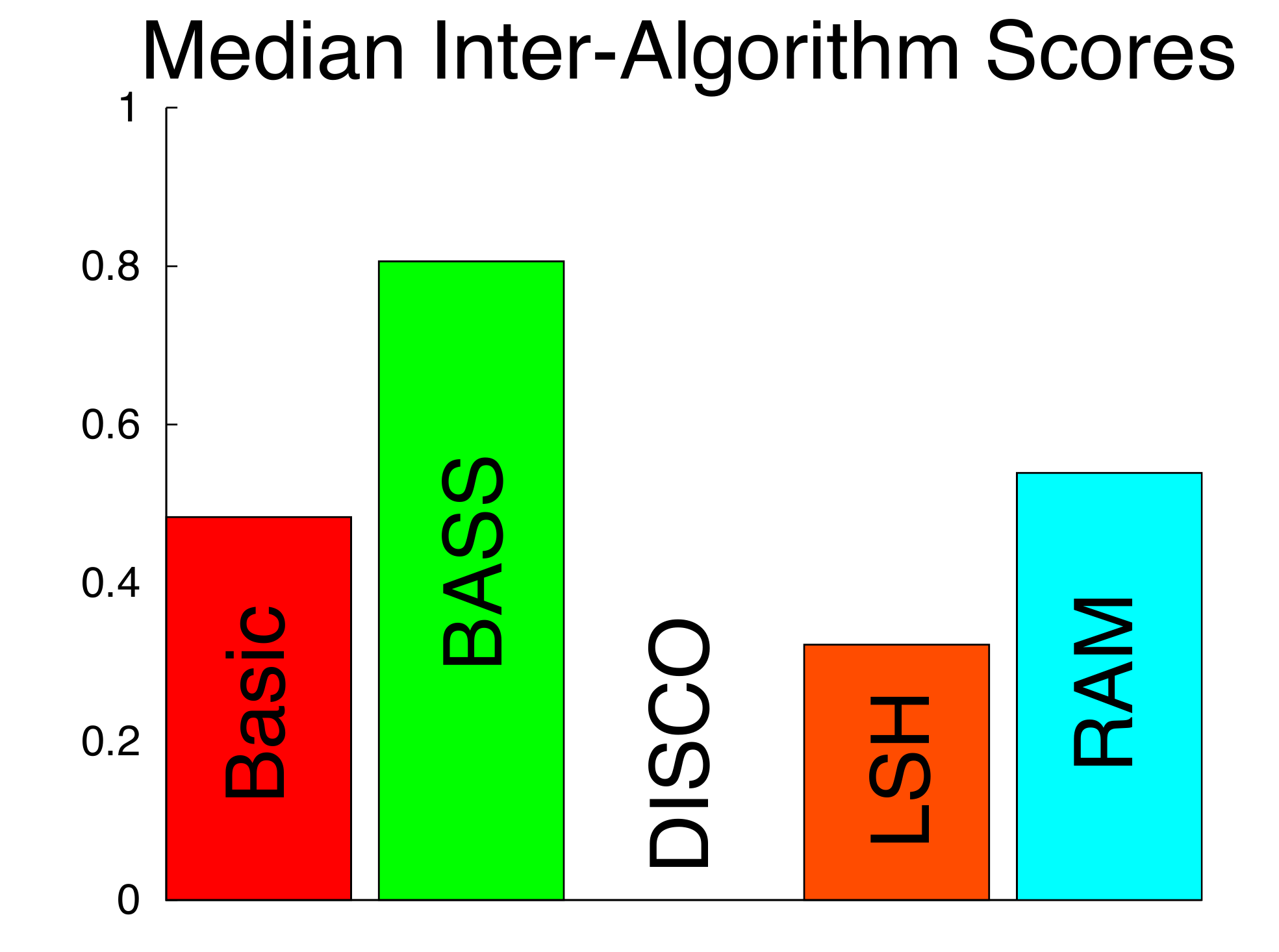}
\caption{
Aggregate normalized scores for the five reinforcement learning agents.
\label{fig:evaluation:aggregate_scores}}
\end{center}
\end{figure}

\subsubsection{Average Score}

The most straightforward method of aggregating normalized scores is to compute their average. Without perfect score normalization, however, score averages tend to be heavily influenced by games such as \gamename{Zaxxon} for which baseline scores are high. Averaging inter-algorithm scores obviates this issue as all 
scores are bounded between 0 and 1. Figure \ref{fig:evaluation:aggregate_scores} displays average 
baseline and inter-algorithm scores for our learning agents.

\subsubsection{Median Score}

Median scores are generally more robust to outliers than average scores. The median is obtained by sorting all normalized scores and selecting the middle element (the average of the two middle elements is used if the number of scores is even). Figure \ref{fig:evaluation:aggregate_scores} shows median baseline and inter-algorithm scores for our learning agents.  Comparing medians and averages in the baseline score (upper two graphs) illustrates exactly the outlier sensitivity of the average score, where the LSH method appears dramatically superior due entirely to its performance in \gamename{Zaxxon}.  

\subsubsection{Score Distribution}

\begin{figure}
\begin{center}
\includegraphics[width=2.95in]{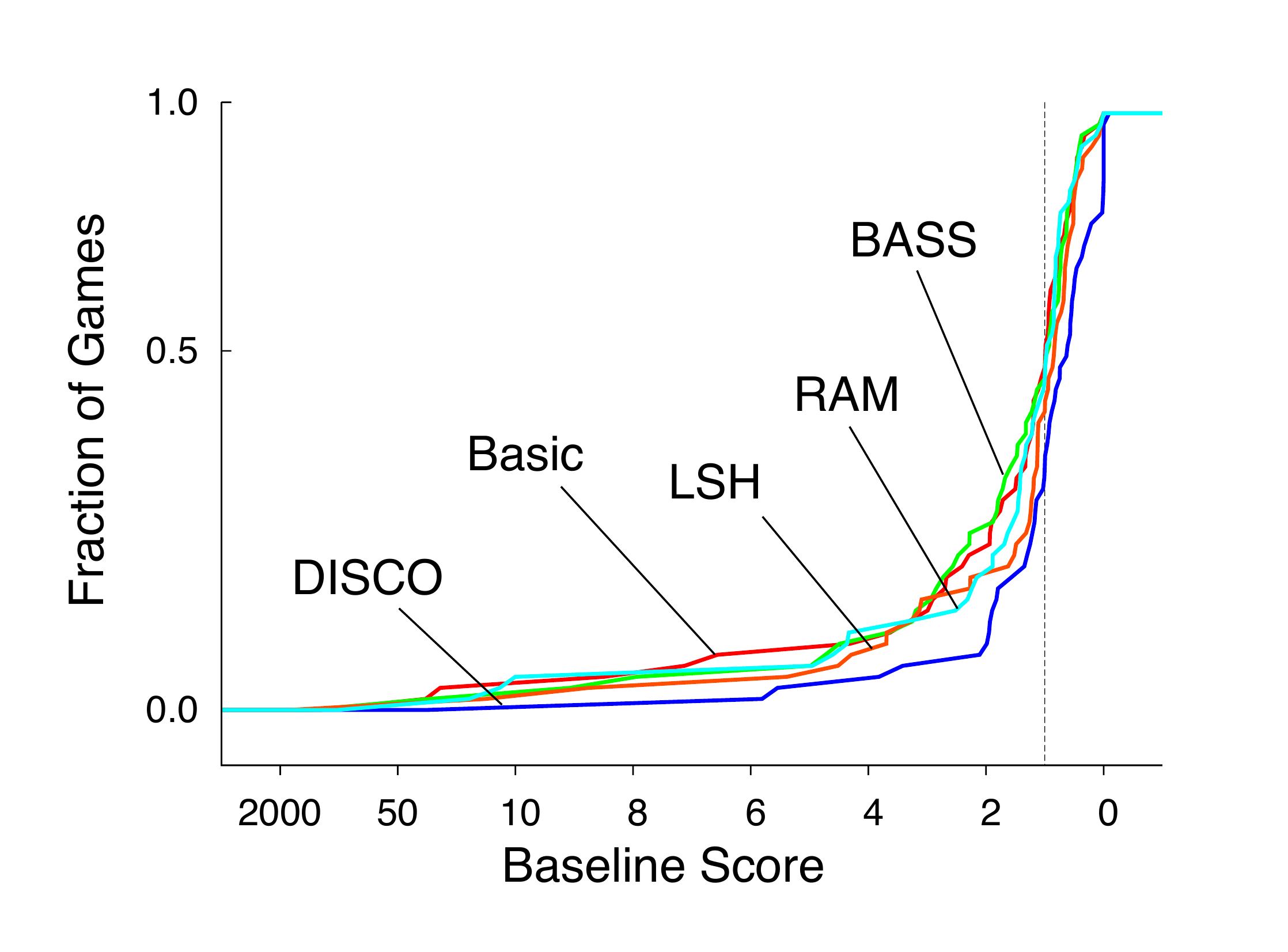}
\vspace{-2ex}
\includegraphics[width=2.95in]{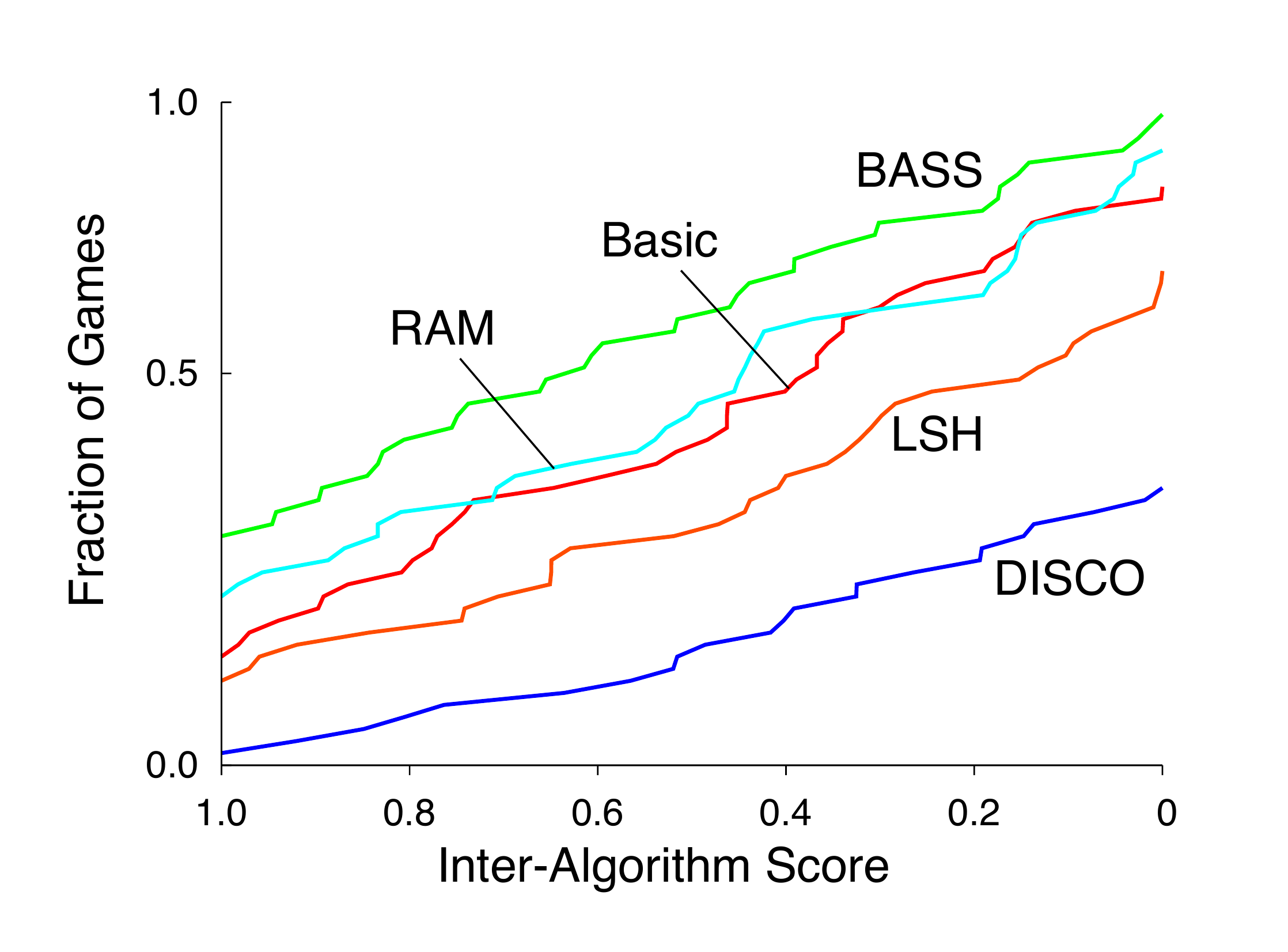}
\caption{Score distribution over all games.\label{fig:evaluation:score_distribution}}
\end{center}
\vspace{-4ex}
\end{figure}

The \emph{score distribution} aggregate is a natural generalization of the median score: it shows the fraction of games on which an algorithm achieves a certain normalized score or better.  It is essentially a quantile plot or inverse empirical CDF. Unlike the average and median scores, the score distribution accurately represents the performance of an agent irrespective of how individual scores are distributed. 
Figure \ref{fig:evaluation:score_distribution} shows baseline and inter-algorithm score distributions. 
Score distributions allow us to compare different algorithms at a glance -- if one curve is above another, the corresponding method generally obtains higher scores. 

Using the baseline score distribution, we can easily determine the proportion of games for which methods perform better than the baseline policies (scores above 1). The inter-algorithm score distribution, on the other hand, effectively conveys the relative performance of each method. 
In particular, it allows us to conclude that BASS performs slightly better than Basic and RAM, and that DISCO performs significantly worse than the other methods.

\begin{table}
\begin{center}
\addtolength{\tabcolsep}{-0.8pt}
\small{
\begin{tabular}{@{}lccccc@{}}
\toprule
& Basic & BASS & DISCO & LSH & RAM \\
\midrule
% Left = first
Basic & --- & 18--32 & 39--13 & 34--18 & 22--25 \\ 
\hline
BASS & 32--18 & --- & 48--5 & 36--17 & 29--20 \\ 
\hline
DISCO & 13--39 & 5--48 & --- & 17--33 & 9--41 \\ 
\hline
LSH & 18--34 & 17--36 & 33--17 & --- & 15--36 \\ 
\hline
RAM & 25--22 & 20--29 & 41--9 & 36--15 & --- \\ 
% Top = first
\bottomrule
\vspace{-1.6em}
\end{tabular}
}
\caption{Paired tests over all games. Each entry shows the number of games for which the performance of the first algorithm (left) is better (--worse) than the second algorithm's.\label{table:paired_tests}}
\end{center}
\vspace{-2em}
\end{table}

\subsection{Paired Tests}

An alternate evaluation metric, especially useful when comparing only a few algorithms, is to perform paired tests over the raw scores. For each game, we performed a two-tailed Welsh's $t$-test with 99\% confidence intervals to determine whether one algorithm's score was statistically different than the other's. Table \ref{table:paired_tests} provides, for each pair of algorithms, the number of games for which one algorithm performs statistically better or worse than the other. Because of their ternary nature, paired tests tend to magnify small but significant differences in scores. 

\section{Related Work}
\label{sec:relatedwork}

We now briefly survey recent research related to Atari 2600 games and some prior work on the construction of empirical benchmarks for measuring general competency.

\subsection{Atari Games}\label{sec:prior_work_on_atari}

There has been some attention devoted to Atari 2600 game playing within the reinforcement learning community.
For the most part, prior work has focused on the challenge of finding good state features for this domain. 
\citeA{diuk2008} applied their DOORMAX algorithm to a restricted version of the game of \gamename{Pitfall!}.
Their method extracts objects from the displayed image with game-specific object detection. 
These objects are then converted into a first-order logic representation of the world, the Object-Oriented Markov Decision Process (OO-MDP). 
Their results show that DOORMAX can discover the optimal behaviour for this OO-MDP within one episode.
\citeA{wintermute2010} proposed a method that also extracts objects from the displayed image and embeds them into a logic-based architecture, SOAR. 
Their method uses a forward model of the scene to improve the performance of the Q-Learning algorithm \cite{watkins1992}. 
They showed that by using such a model, a reinforcement learning agent could learn to play a restricted version of the game of \gamename{Frogger}.
\citeA{cobo2011} investigated automatic feature discovery in the games of \gamename{Pong} and \gamename{Frogger}, using their own simulator. 
Their proposed method takes advantage of human trajectories to identify state features that are important for playing console games. 
Recently, \citeA{hausknecht_12} proposed HyperNEAT-GGP, an evolutionary approach for finding policies to play Atari 2600 games. 
Although HyperNEAT-GGP is presented as a general game playing approach, it is currently difficult to assess its general performance as the reported results were limited to only two games. 
Finally, some of the authors of this paper \cite{bellemare2012} recently presented a domain-independent feature generation technique that attempts to focus its effort around the location of the player avatar.
This work used the evaluation methodology advocated here and is the only one to demonstrate the technique across a large set of testing games.

\subsection{Evaluation Frameworks for General Agents}

Although the idea of using games to evaluate the performance of agents has a long history in artificial intelligence, it is only more recently that an emphasis on generality has assumed a more prominent role.
\citeA{pell93strategy} advocated the design of agents that, given an abstract description of a game, could automatically play them.
His work strongly influenced the design of the now annual General Game Playing competition \cite{ggpcomp}.
Our framework differs in that we do not assume to have access to a compact logical description of the game semantics.
\citeA{schaul11} also recently presented an interesting proposal for using games to measure the general capabilities of an agent.
\citeA{whiteson11} discuss a number of challenges in designing empirical tests to measure general reinforcement learning performance; this work can be seen as attempting to address their important concerns.

Starting in 2004 as a conference workshop, the Reinforcement Learning competition \cite{whiteson10} was held until 2009 (a new iteration of the competition has been announced for 2013\footnote{\url{http://www.rl-competition.org}}). Each year new domains are proposed, including standard RL benchmarks, Tetris, and Infinite Mario \cite{mohan_laird_09}. In a typical competition domain, the agent's state information is summarized through a series of high-level state variables rather than direct sensory information. Infinite Mario, for example, provides the agent with an object-oriented observation space. In the past, organizers have provided a special `Polyathlon' track in which agents must behave in a medley of continuous-observation, discrete-action domains.

Another longstanding competition, the International Planning Competition (IPC)\footnote{\url{http://ipc.icaps-conference.org}}, has been organized since 1998, and aims to ``produce new benchmarks, and to gather and disseminate data about the current state-of-the-art'' \cite{coles_12}. The IPC is composed of different tracks corresponding to different types of planning problems, including factory optimization, elevator control and agent coordination. For example, one of the problems in the 2011 competition consists in coordinating a set of robots around a two-dimensional gridworld so that every tile is painted with a specific colour. Domains are described using either relational reinforcement learning, yielding parametrized Markov Decision Processes (MDPs) and Partially Observable MDPs, or using logic predicates, e.g. in STRIPS notation. 

One indication of how much these competitions value domain variety can be seen in the time spent on finding a good specification language. The 2008-2009 RL competitions, for example, used RL-Glue\footnote{\url{http://glue.rl-community.org}} specifically for this purpose; the 2011 planning under uncertainty track of the IPC similar employed the Relation Dynamic Influence Diagram Language. While competitions seek to spur new research and evaluate existing algorithms through a standardized set of benchmarks, they are \emph{not} independently developed, in the sense that the vast majority of domains are provided by the research community. Thus a typical competition domain reflects existing research directions: Mountain Car and Acrobot remain staples of the RL competition. These competitions also focus their research effort on domains that provide high-level state variables, for example the location of robots in the floor-painting domain described above. By contrast, the Arcade Learning Environment and the domain-independent setting force us to consider the question of perceptual grounding: how to extract meaningful state information from raw game screens (or RAM information). In turn, this emphasizes the design of algorithms that can be applied to sensor-rich domains without significant expert knowledge. 

There have also been a number of attempts to define formal agent performance metrics based on algorithmic information theory.
The first such attempts were due to \citeA{Hernandez-orallo98aformal} and to \citeA{DoweHajek98}.
More recently, the approaches of \citeA{Hern10} and of \citeA{legg11} appear to have some potential.
Although these frameworks are general and conceptually clean, the key challenge remains how to specify sufficiently interesting classes of environments.
In our opinion, much more work is required before these approaches can claim to rival the practicality of using a large set of existing human-designed environments for agent evaluation.

\section{Final Remarks}
\label{sec:final-remarks}

The Atari 2600 games were developed for humans and as such exhibit many idiosyncrasies that make them both challenging and exciting. Consider, for example, the game \gamename{Pong}. \gamename{Pong} has been studied in a variety of contexts as an interesting reinforcement learning domain \cite{cobo2011,stober08pixels,monroy06coevolution}. The Atari 2600 \gamename{Pong}, however, is significantly more complex than \gamename{Pong} domains developed for research. Games can easily last 10,000 time steps (compared to 200--1000 in other domains); observations are composed of 7-bit $160 \times 210$ images (compared to $300 \times 200$ black and white images in the work of \citeA{stober08pixels}, or 5-6 input features elsewhere); observations are also more complex, containing the two players' score and side walls. In sheer size, the Atari 2600 \gamename{Pong} is thus a larger domain. Its dynamics are also more complicated.  In research implementations of \gamename{Pong} object motion is implemented using first-order mechanics.  However, in Atari 2600 \gamename{Pong} paddle control is nonlinear: simple experimentation shows that fully predicting the player's paddle requires knowledge of the last 18 actions. As with many other Atari games, the player paddle also moves every other frame, adding a degree of temporal aliasing to the domain. 

While Atari 2600 \gamename{Pong} may appear unnecessarily contrived, it in fact reflects the unexpected complexity of the problems with which humans are faced. Most, if not all Atari 2600 games are subject to similar programming artifacts: in \gamename{Space Invaders}, for example, the invaders' velocity increases nonlinearly with the number of remaining invaders. In this way the Atari 2600 platform provides AI researchers with something unique: clean, easily-emulated domains which nevertheless provide many of the challenges typically associated with real-world applications.

Should technology advance so as to render general Atari 2600 game playing achievable, our challenge problem can always be extended to use more recent video game platforms. A natural progression, for example, would be to move on to the Commodore 64, then to the Nintendo, and so forth towards current generation consoles.
All of these consoles have hundreds of released games, and older platforms have readily available emulators.  With the ultra-realism of current generation consoles, each console represents a natural stepping stone toward general real-world competency.  Our hope is that by using the methodology advocated in this paper, we can work in a bottom-up fashion towards developing more sophisticated AI technology while still maintaining empirical rigor.  

\section{Conclusion}
\label{sec:conclusion}

This article has introduced the Arcade Learning Environment, a platform for evaluating the development of general, domain-independent agents.  
ALE provides an interface to hundreds of Atari 2600 game environments, each one different, interesting, and designed to be a challenge for human players.  
We illustrate the promise of ALE as a challenge problem by benchmarking several domain-independent agents that use well-established reinforcement learning and planning techniques.  
Our results suggest that general Atari game playing is a challenging but not intractable problem domain with the potential to aid the development and evaluation of general agents.

\acks
We would like to thank Marc Lanctot, Erik Talvitie, and Matthew Hausknecht for providing suggestions on helping debug and improving the Arcade Learning Environment source code. We would also like to thank our reviewers for their helpful feedback and enthusiasm about the Atari 2600 as a research platform. The work presented here was supported by the Alberta Innovates Technology Futures, the Alberta Innovates Centre for Machine Learning at the University of Alberta, and the Natural Science and Engineering Research Council of Canada. Invaluable computational resources were provided by Compute/Calcul Canada.

\newpage
\appendix

\section{Feature Set Construction}\label{apdx:feature_sets}

This section gives a detailed description of the five feature generation techniques from Section \ref{sec:RL}. 

\subsection{Basic Abstraction of the ScreenShots (BASS)}
\label{sec:agents:rl:bass}
\begin{figure}[h!]
\begin{center}
\includegraphics[width=2.5in]{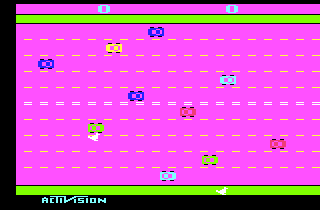}
\includegraphics[width=2.5in]{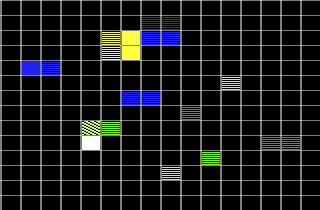}
\caption{\textbf{Left:} Freeway in SECAM colours. \textbf{Right:} BASS colour encoding for the same screen. \label{fig:agents:rl:bass:features}}
\end{center}
\end{figure}

The idea behind BASS is to directly encode colours present on the screen. This method is motivated by three observations on the Atari 2600 hardware and games:
\begin{enumerate}
	\item While the Atari 2600 hardware supports a screen resolution of $160 \times 210$, game objects are usually larger than a few pixels. Overall, important game events happen at a much lower resolution.
	\item Many Atari 2600 games have a static background, with a few important objects moving on the screen. While the screen matrix is densely populated, the actual interesting features on the screen are often sparse.  
	\item While the hardware can show up to 128 colours in the NTSC mode, it is limited to only 8 colours in the SECAM mode. Consequently, most games use a few number of colours to distinguish important objects on the screen. 
\end{enumerate}
The game screen is first preprocessed by subtracting its background, detected using a simple histogram method. BASS then encodes the presence of each of the eight SECAM palette colours at a low resolution, as depicted in Figure \ref{fig:agents:rl:bass:features}. Intuitively, BASS seeks to capture the presence of objects of certain colours at different screen locations. BASS also encodes relations between objects by constructing all pairwise combinations of its encoded colour features. In \gamename{Asterix}, for example, it is important to know if there is a green object (player character) \emph{and} a red object (collectable object) in its vicinity. Pairwise features allow us to capture such object relations.

\subsection{Basic}
\label{sec:agents:rl:basic}

The Basic method generates the same set of features as BASS, but omits the pairwise combinations. This allows us to study whether the additional features are beneficial or harmful to learning. Because the Basic method has fewer features than BASS, it encodes the presence of each of the 128 colours. In comparison to BASS, Basic therefore represents colour more accurately, but cannot represent object interactions. 

\subsection{Detecting Instances of Classes of Objects (DISCO)}
\label{sec:agents:rl:disco}
\begin{figure}[htb]
\begin{center}
\includegraphics[width=2.5in]{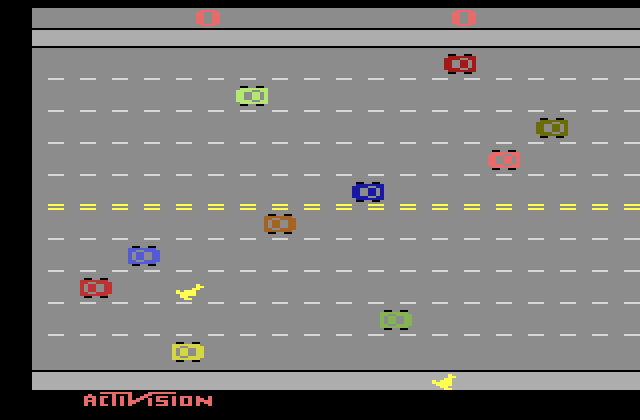}
\includegraphics[width=2.5in]{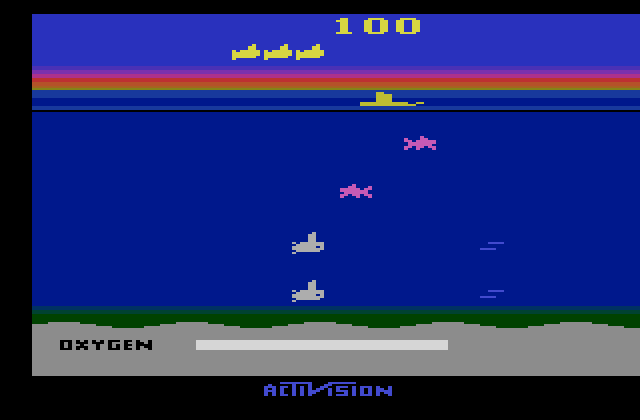}
\caption{\textbf{Left:} Screenshot of the game \gamename{Freeway}. Although there are ten different cars, they can all be considered as instances of a single class. \textbf{Right:} Screenshot of the game \gamename{Seaquest} depicting four different object classes.\label{fig:agents:rl:disco:screenshot}}
\end{center}
\end{figure}

This feature generation method is based on detecting a set of classes representing game entities and locating instances of these classes on the screen. DISCO is motivated by the following additional observations on Atari 2600 games:
\begin{enumerate}
	\item The game entities are often instances of a few \emph{classes} of objects. For instance, as Figure \ref{fig:agents:rl:disco:screenshot} shows, while there are many objects in a sample screen of the game \gamename{Freeway}, all of these objects are instances of only two classes: \emph{Chicken} and \emph{Car}. Similarly, all the objects on a sample screen of the game \gamename{Seaquest} are instances of one of these six classes: \emph{Fish, Swimmer, Player Submarine, Enemy Submarine, Player Bullet,} and \emph{Enemy Bullet}. 
	\item The interaction between two objects can often be generalized to all instances of their respective classes. As an example, consider \emph{Car}-\emph{Chicken} object interactions in \gamename{Freeway}: learning that there is lower value associated with one \emph{Chicken} instance hitting a \emph{Car} instance can be generalized to all instances of those two classes. 
\end{enumerate}

DISCO first performs a series of preprocessing steps to discover classes,
during which no value function learning is performed. When the agent subsequently learns 
to play the game, DISCO generates features by detecting objects on the screen and classifying them. 
The DISCO process is summarized by the following steps:

\begin{algorithm}
\caption{Locally Sensitive Hashing (LSH) Feature Generation\label{algo:lsh_feature_generation}}
\begin{algorithmic}
\item[\textbf{Constants.}] $M$ (hash table size), $n$ (screen bit vector size) \\
  $l$ (number of random bit vectors), $k$ (number of non-zero entries) 
\item[\textbf{Initialization (once).}]
\STATE $\{v_1 \dots v_l \} \gets \text{generateRandomVectors}(l, k, n)$ 
\STATE $\{\text{hash}_1 \dots \text{hash}_l \} \gets \text{generateHashFunctions}(l, M, n)$ 
\STATE
\item[\textbf{Input.}] A screen matrix $I$ with elements $I_{xy} \in \{0, \dots, 127 \}$
\STATE

\item[$\text{LSH}(I)$]
\STATE $s \gets \textrm{binarizeScreen}(I)$ ($s$ has length $n$)
\STATE Initialize $\phi \in \bR^{lM} = 0$ 
\FOR{$i = 1 \dots l$} 
  \STATE $h = 0$
  \FOR{$j = 1 \dots n$}
    \STATE $h \gets h + \indic{s_j = v_{ij}} \text{hash}_i[j] \mod M$ \COMMENT{hash the projection of $s$ onto $v_i$} 
  \ENDFOR
  \STATE $\phi[M (i - 1) + h] = 1$ \COMMENT{one binary feature per random bit vector}
\ENDFOR
\STATE
\item[$\text{binarizeScreen}(I)$]
\STATE Initialize $s \in \bR^n = 0$
\FOR{$y = 1 \dots h$, $x = 1 \dots w$ ($h = 210, w = 160$)}
  \STATE $s[x + y * h + I_{xy}] = 1$
\ENDFOR
\RETURN $s$
\STATE

\item[$\text{generateRandomVectors}(l, k, n)$]
\STATE Initialize $v_1 \dots v_l \in \bR^n = 0$
\FOR{$i = 1 \dots l$}
  \STATE Select $x_1, x_2, \dots, x_k$ distinct coordinates between 1 and $n$ uniformly at random 
  \STATE $v_i[x_1] = 1$; $v_i[x_2] = 1$; \dots ; $v_i[x_k] = 1$
\ENDFOR
\RETURN $\{ v_1, \dots v_l \}$
\STATE

\item[$\text{generateHashFunctions}(l, M, n)$ (hash functions are vectors of random coordinates)]
\STATE Initialize $\text{hash}_1 \dots \text{hash}_l \in \bR^n = 0$
\FOR{$i = 1 \dots l$, $j = 1 \dots n$}
  \STATE $\text{hash}_i[j] \gets \text{random}(1, M)$ (uniformly random coordinate between 1 and M)
\ENDFOR
\RETURN $\{ \text{hash}_1, \dots \text{hash}_l \}$
\STATE
\item[\textbf{Remark.} With sparse vector operations, LSH has a $O(lk + n)$ cost per step.] 

\end{algorithmic}
\end{algorithm}

\begin{itemize}
	\item Preprocessing: 
	\begin{itemize}
		\item \textbf{Background detection:} The static background matrix is extracted using a histogram method, as with BASS. 
		\item \textbf{Blob extraction:} A list of moving blob (foreground) objects is detected in each game screen. 
		\item \textbf{Class discovery:} A set of classes is detected from the extracted blob objects.
    \item \textbf{Class filtering:} Classes that appear infrequently or are restricted to small region of the screen are removed from the set.
    \item \textbf{Class merging:} Classes that have similar shapes are merged together.
	\end {itemize}
	\item Feature generation: 
	\begin{itemize}
		\item \textbf{Class instance detection:} At each time step, class instances are detected from the current screen matrix.
		\item \textbf{Feature vector generation:}  A feature vector is generated from the detected instances by tile-coding their absolute position as well as the relative position and velocity of every pair of instances from different classes. Multiple instances of the same objects are combined additively. 
	\end {itemize}
\end{itemize}

\begin{figure}
\begin{center}
\includegraphics[width=2.5in]{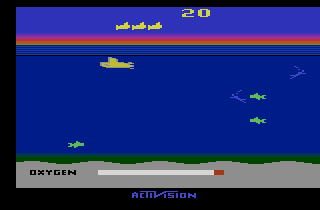}
\includegraphics[width=2.5in]{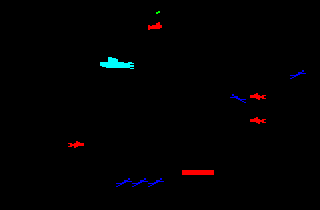}
\caption{\textbf{Left:} Screenshot of the game \gamename{Seaquest}. \textbf{Right:} Objects detected by DISCO in the game Seaquest. Each colour represents a different class.\label{fig:agents:rl:disco:features}}
\end{center}
\end{figure}

Figure \ref{fig:agents:rl:disco:features} shows discovered objects in a Seaquest frame. This image illustrates the difficulties in detecting objects: although DISCO correctly classifies the different fish as part of the same class, it also detects a life icon and the oxygen bar as part of that class. 

\subsection{Locality Sensitive Hashing (LSH)}

An alternative approach to BASS and DISCO is to use well-established feature generation methods that are agnostic about the type of input they receive. Such methods include polynomial bases \cite{schweitzer_85}, sparse distributed memories \cite{kanerva_88} and locality sensitive hashing (LSH) \cite{gionis_99}. In this paper we consider the latter as a simple mean of reducing the large image space to a smaller, more manageable set of features. The input -- here, a game screen -- is first mapped to a bit vector of size $7 \times 210 \times 160$. The resulting vector is then hashed down into a smaller set of features. LSH performs an additional random projection step to ensure that similar screens are more likely to be binned together. The LSH generation method is detailed in Algorithm \ref{algo:lsh_feature_generation}. 
\subsection{RAM-based Feature Generation}
\label{sec:agents:rl:ram}
Unlike the previous three methods, which generate feature vectors based on the game screen, the RAM-based feature generation method relies on the contents of the console memory. The Atari 2600 has only $128 \times 8 = 1024$ bits of random access memory\footnote{Some games provided more RAM on the game cartridge: the \emph{Atari Super Chip}, for example, offered an additional 128 bytes of memory. The current approach only considers the main memory included in the Atari 2600 console.}, which must hold the complete internal state of a game: location of game entities, timers, health indicators, etc. The RAM is therefore a relatively compact representation of the game state, and in contrast to the game screen, it is also Markovian. The purpose of our RAM-based agent is to investigate whether features generated from the RAM affect performance differently from features generated from game screens. 

The first part of the generated feature vector simply includes the 1024 bits of RAM. Atari 2600 game programmers often used these bits not as individual values, but as part of 4-bit or 8-bit words. Linear function approximation on the individual bits can capture the value of these multi-bit words. We are also interested in the relation between pairs of values in memory. To capture these relations, the logical-AND of all possible bit pairs is appended to the feature vector. Note that a linear function on the pairwise $AND$'s can capture products of both 4-bit and 8-bit words. This is because the product of two $n$-bit words can be expressed as a weighted sum of the pairwise products of their bits.

\newpage
\section{UCT Pseudocode}\label{appendix:uct_pseudocode}

\begin{algorithm}
\caption{UCT\label{algo:uct_algorithm}}
\begin{algorithmic}
\item[\textbf{Constants.}] $m$ (search horizon), $k$ (simulations per step)
\item[\textbf{Variables.}] $\Psi$ (search tree)
\item[\textbf{Input.}] $s$ (current state)
\STATE

\item[$\text{UCT}(s)$]
\IF{$\Psi$ is empty or $\text{root}(\Psi) \ne s$}
  \STATE $\Psi \gets \text{empty search tree}$
  \STATE $\text{root}(\Psi) \gets s$
\ENDIF
\REPEAT
  \STATE $\text{sample}(\Psi, m)$
\UNTIL{$\text{visits}(\text{root}(\Psi)) = k$}
\STATE $a \gets \text{bestAction}(\Psi)$
\STATE $\text{prune}(\Psi,a)$ \COMMENT{optional}
\RETURN $a$

\STATE
\item[$\text{sample}(\Psi, m)$]
\STATE{$n \gets \text{root}(\Psi)$}
\WHILE{$n$ is not a leaf, $m > \text{depth}(n)$}
  \IF{some action $a$ was never taken in $n$}
    \STATE $(c,\text{reward}) \gets \text{emulate}(n,a)$ \COMMENT{run model for one step} 
    \STATE $\text{immediate-return}(c) \gets \text{reward}$
    \STATE $\text{child}(n,a) \gets c$
    \STATE $n \gets c$ \COMMENT{c is necessarily a leaf} 
  \ELSE
    \STATE $a \gets \text{selectAction}(n)$
    \STATE $n \gets \text{child}(n, a)$ 
  \ENDIF
\ENDWHILE
\STATE $R = \text{rollout}(n, m - depth(n))$
\STATE $\text{update-value(n, R)}$ \COMMENT{propagate values back up}

\STATE
\item[$\text{bestAction}(\Psi)$]
\RETURN $\arg \max_a \left [ \text{visits}(\text{child}(\text{root}(\Psi), a)) \right ]$ \COMMENT{action most frequently taken at root}

\STATE
\item[$\text{prune}(\Psi, a)$]
\STATE $\text{root}(\Psi) \gets \text{child}(\text{root}(\Psi), a)$

\end{algorithmic}
\end{algorithm}

\begin{algorithm}
\caption{UCT Routines\label{algo:uct_routines}}
\begin{algorithmic}
\item[\textbf{Constants.}] $\gamma$ : discount factor

\STATE
\item[$\text{selectAction}(n)$]
\FORALL{$c$ children of $n$}
  \STATE $V(c) \gets \text{average-return}(c) + \sqrt{\frac{\log\left [ {\text{visits}(c)}  \right ]}{\text{visits}(n)}}$
\ENDFOR
\RETURN $\arg \max_a V(\text{child}(n,a))$

\STATE
\item[$\text{rollout}(n, m)$]
\STATE $R = 0$ \COMMENT{Initialize Monte-Carlo return to 0}
\STATE $g = 1$
\WHILE{$m > 0$}
  \STATE Select $a$ according to some rollout policy (e.g. uniformly randomly)
  \STATE $(n, \text{reward}) \gets \text{emulate}(n, a)$
  \STATE $R \gets R + g \times \text{reward}$
  \STATE $m \gets m - 1$
  \STATE $g \gets g \times \gamma$
\ENDWHILE
\RETURN $R$ 

\STATE
\item[$\text{update-value}(n, R)$]
\STATE $R \gets R + \text{immediate-reward(n)}$ 
\STATE $\text{average-return}(n) \gets \text{average-return}(n) \frac{\text{visits}(n)}{\text{visits}(n)+1} + \frac{R}{\text{visits}(n) + 1}$
\STATE $\text{visits}(n) \gets \text{visits}(n) + 1$
\IF{$n$ is not the root of $\Psi$, i.e. $\text{parent}(n) \ne \text{null}$}
  \STATE $\text{update-value}(\text{parent}(n), \gamma \times R)$
\ENDIF
\end{algorithmic}
\end{algorithm}

\newpage
\section{Experimental Parameters}
\label{appendix:parameters}

\begin{table}[h!]
\small
\begin{center}
\begin{tabular}{|r||r||r|l|}
\hline
\textbf{General} & All experiments & Maximum frames per episode & 18,000 \\
\cline{3-4}
& & Frames per action & 5 \\
\cline{2-4}
& Reinforcement learning & Training episodes per trial & 5,000 \\
\cline{3-4}
& & Evaluation episodes per trial & 500 \\
\cline{3-4}
& & Number of trials per result & 30 \\ 
\hline
\hline
\textbf{Preprocessing} & Background detection & Sample screens per game & 18,000 \\
\cline{2-4}
& Class discovery & Sample screens per game & 36,000 \\
\cline{3-4}
& & Maximum number of classes & 10 \\
\cline{3-4}
& & Maximum object velocity (pixels) & 8 \\
\cline{3-4}
& & Minimum frequency of class appearance & 20\% \\
\hline
\hline
\textbf{Reinforcement} & All agents & Discount factor $\gamma$ & 0.999 \\
\cline{3-4} 
\textbf{learning} & & Exploration rate $\epsilon$ & 0.05 \\
\cline{2-4} 

& BASS and & Learning rate $\alpha$ & 0.5 \\
\cline{3-4}
& Basic & Eligibility traces decay rate $\lambda$ & 0.9 \\
\cline{3-4}
& & Grid width & 16 \\
\cline{3-4}
& & Grid height & 14 \\
\cline{2-4}
& BASS only & Number of different colours & 8 \\
\cline{2-4}
& Basic only & Number of different colours & 128 \\
\cline{2-4}

& DISCO & Learning rate $\alpha$ & 0.1 \\
\cline{3-4}
& & Eligibility traces decay rate $\lambda$ & 0.9 \\
\cline{3-4}
& & Tile coding, number of tilings & 8 \\ 
\cline{3-4}
& & Tile coding, grid size & 8 \\ 
\cline{2-4}

& RAM-based & Learning rate $\alpha$ & 0.2 \\
\cline{3-4}
& & Eligibility traces decay rate $\lambda$ & 0.5 \\
\cline{2-4}

& LSH & Learning rate $\alpha$ & 0.5 \\
\cline{3-4}
& & Eligibility traces decay rate $\lambda$ & 0.5 \\
\cline{3-4}
& & Number of random vectors $l$ & 2000 \\
\cline{3-4}
& & Number of non-zero vector entries $k$ & 1000 \\
\cline{3-4}
& & Per-vector hash table size $M$ & 50 \\ 
\hline
\hline
\textbf{Planning} & UCT & Simulations per action & 500 \\
\cline{3-4}
& & Maximum search depth (frames) & 300 \\
\cline{3-4}
& & Exploration constant & 0.1 \\
\cline{2-4}
& Full-tree search & Maximum frames emulated per action & 133,000 \\
\hline
\end{tabular}
\end{center}
\end{table}

\newpage
\section{Detailed Results}\label{appendix:detailed_results}

\subsection{Reinforcement Learning}

\begin{table}[h!]
\tiny
\begin{center}
\include{tables/allRLResults}
\caption{Reinforcement Learning results. The first five games constitute our training set. See Section \ref{sec:RL} for details.\label{table:appendix:detailed_results:rl}}.
\end{center}
\end{table}

\newpage
\subsection{Planning}
\begin{table}[h!]
\tiny
\begin{center}
\include{tables/allSearchResults}
\caption{Search results. The first five games constitute our training set. See Section \ref{sec:planning} for details.\label{table:appendix:detailed_results:search}}.
\end{center}
\end{table}

\newpage

\nocite{Russell97rationalityand,Hutter:04uaibook,legg08machine}
\bibliography{atari}
\bibliographystyle{theapa}

\end{document}

%% file: tables/selectedRLResults.tex
\begin{tabular}{|r|r|r|r|r|r||r|r|r|r|}
\hline
Game & Basic & BASS & DISCO & LSH & RAM & Random & Const & Perturb & Human \\
\hline
\hline
\gamename{Asterix} & 862& 860& 755& \textbf { 987 }& 943& 288& 650& 338& 620\\
\hline
\gamename{Seaquest} & 579& \textbf { 665 }& 422& 509& 594& 108& 160& 451& 156\\
\hline
\gamename{Boxing} & -3& 16& 12& 10& \textbf { 44 }& -1& -25& -10& -2\\
\hline
\gamename{H.E.R.O.} & 6053& \textbf { 6459 }& 2720& 3836& 3281& 712& 0& 148& 6087\\
\hline
\gamename{Zaxxon} & 1392& 2069& 70& \textbf { 3365 }& 304& 0& 0& 2& 820\\
\hline
\end{tabular}

%% file: tables/selectedSearchResults.tex
\begin{tabular}{|r|r|r|r|r|}
\hline
Game & Full Tree & UCT & Best Learner & Best Baseline \\ 
\hline
\hline
\gamename{Asterix} & 2136& \textbf { 290700 }& 987& 650\\
\hline
\gamename{Seaquest} & 288& \textbf { 5132 }& 665& 451\\
\hline
\gamename{Boxing} & \textbf { 100 }& 100& 44& -1\\
\hline
\gamename{H.E.R.O.} & 1324& \textbf { 12860 }& 6459& 712\\
\hline
\gamename{Zaxxon} & 0& \textbf { 22610 }& 3365& 2\\
\hline
\end{tabular}

%% file: tables/allRLResults.tex
\begin{tabular}{|r|r|r|r|r|r||r|r|r|}
%{|c|c|c|c|c|c||c|c|c|}
\hline
Game & Basic & BASS & DISCO & LSH & RAM & Random & Const & Perturb \\
\hline
\hline
\gamename{Asterix} & 862.3& 859.8& 754.6& \textbf { 987.3 }& 943.0& 288.1& 650.0& 337.8\\
\hline
\gamename{Beam Rider} & 929.4& 872.7& 563.0& 793.6& 729.8& 434.7& \textbf { 996.0 }& 754.8\\
\hline
\gamename{Freeway} & 11.3& 16.4& 12.8& 15.4& 19.1& 0.0 & 21.0& \textbf { 22.5 }\\
\hline
\gamename{Seaquest} & 579.0& \textbf { 664.8 }& 421.9& 508.5& 593.7& 107.9& 160.0& 451.1\\
\hline
\gamename{Space Invaders} & 203.6& 250.1& 239.1& 222.2& 226.5& 156.1& 245.0& \textbf { 270.5 }\\
\hline
\hline
\gamename{Alien} & \textbf { 939.2 }& 893.4& 623.6& 510.2& 726.4& 102.0& 140.0& 313.9\\
\hline
\gamename{Amidar} & 64.9& \textbf { 103.4 }& 67.9& 45.1& 71.4& 0.8& 31.0& 37.8\\
\hline
\gamename{Assault} & 465.8& 378.4& 371.7& \textbf { 628.0 }& 383.6& 334.3& 357.0& 497.8\\
\hline
\gamename{Asteroids} & 829.7& 800.3& 744.5& 590.7& 907.3& \textbf { 1526.7 }& 140.0& 539.9\\
\hline
\gamename{Atlantis} & \textbf { 62687.0 }& 25375.0& 20857.3& 17593.9& 19932.7& 33058.4& 1500.0& 12089.1\\
\hline
\gamename{Bank Heist} & 98.8& 71.1& 51.4& 64.6& \textbf { 190.8 }& 15.0& 0.0& 13.5\\
\hline
\gamename{Battle Zone} & 15534.3& 12750.8& 0.0& 14548.1& \textbf { 15819.7 }& 2920.0& 13000.0& 5772.0\\
\hline
\gamename{Berzerk} & 329.2& 491.3& 329.0& 441.0& 501.3& 233.8& \textbf { 670.0 }& 552.9\\
\hline
\gamename{Bowling} & 28.5& \textbf { 43.9 }& 35.2& 26.1& 29.3& 24.6& 30.0& 30.0\\
\hline
\gamename{Boxing} & -2.8& 15.5& 12.4& 10.5& \textbf { 44.0 }& -1.5& -25.0& -10.1\\
\hline
\gamename{Breakout} & 3.3& \textbf { 5.2 }& 3.9& 2.5& 4.0& 1.5& 3.0& 2.9\\
\hline
\gamename{Carnival} & \textbf { 2323.9 }& 1574.2& 1646.3& 1147.2& 765.4& 869.2& 0.0& 485.4\\
\hline
\gamename{Centipede} & 7725.5& 8803.8& 6210.6& 6161.6& 7555.4& 2805.1& \textbf { 16527.0 }& 8937.2\\
\hline
\gamename{Chopper Command} & 1191.4& \textbf { 1581.5 }& 1349.0& 943.0& 1397.8& 698.2& 1000.0& 973.7\\
\hline
\gamename{Crazy Climber} & 6303.1& 7455.6& 4552.9& 20453.7& \textbf { 23410.6 }& 2335.4& 0.0& 2235.0\\
\hline
\gamename{Demon Attack} & 520.5& 318.5& 208.8& 355.8& 324.8& 289.3& 130.0& \textbf { 776.2 }\\
\hline
\gamename{Double Dunk} & -15.8& -13.1& -23.2& -21.6& -20.3& -15.6& \textbf { 0.0 }& -20.3\\
\hline
\gamename{Elevator Action} & 3025.2& 2377.6& 4.6& \textbf { 3220.6 }& 507.9& 1040.9& 0.0& 562.9\\
\hline
\gamename{Enduro} & 111.8& \textbf { 129.1 }& 0.0& 95.8& 112.3& 0.0& 9.0& 25.9\\
\hline
\gamename{Fishing Derby} & -92.6& -92.1& \textbf { -89.5 }& -93.2& -91.6& -93.8& -99.0& -97.2\\
\hline
\gamename{Frostbite} & 161.0& 161.1& 176.6& \textbf { 216.9 }& 147.9& 70.3& 160.0& 175.2\\
\hline
\gamename{Gopher} & 545.8& \textbf { 1288.3 }& 295.7& 941.8& 722.5& 243.7& 0.0& 286.8\\
\hline
\gamename{Gravitar} & 185.3& 251.1& 197.4& 105.9& \textbf { 387.7 }& 205.4& 0.0& 106.0\\
\hline
\gamename{H.E.R.O.} & 6053.1& \textbf { 6458.8 }& 2719.8& 3835.8& 3281.1& 712.0& 0.0& 147.5\\
\hline
\gamename{Ice Hockey} & -13.9& -14.8& -18.9& -15.1& -9.5& -14.8& \textbf { -1.0 }& -6.5\\
\hline
\gamename{James Bond} & 197.3& \textbf { 202.8 }& 17.3& 77.1& 133.8& 23.3& 0.0& 82.0\\
\hline
\gamename{Journey Escape} & -8441.0& -14730.7& -9392.2& -13898.9& -8713.5& -18201.7& \textbf { 0.0 }& -10693.9\\
\hline
\gamename{Kangaroo} & 962.4& \textbf { 1622.1 }& 457.9& 256.4& 481.7& 44.4& 200.0& 498.4\\
\hline
\gamename{Krull} & 2823.3& \textbf { 3371.5 }& 2350.9& 2798.1& 2901.3& 1880.1& 0.0& 1690.1\\
\hline
\gamename{Kung-Fu Master} & 16416.2& \textbf { 19544.0 }& 3207.0& 8715.6& 10361.1& 488.2& 0.0& 578.4\\
\hline
\gamename{Montezuma's Revenge} & \textbf { 10.7 }& 0.1& 0.0& 0.1& 0.3& 0.3& 0.0& 0.0\\
\hline
\gamename{Ms. Pac-Man} & 1537.2& \textbf { 1691.8 }& 999.6& 1070.8& 1021.1& 163.3& 210.0& 505.5\\
\hline
\gamename{Name This Game} & 1818.9& 2386.8& 1951.0& 2029.8& 2500.1& 2012.3& \textbf { 3080.0 }& 1854.3\\
\hline
\gamename{Pooyan} & 800.3& 1018.9& 402.7& \textbf { 1225.3 }& 1210.9& 501.1& 30.0& 540.8\\
\hline
\gamename{Pong} & -19.2& \textbf { -19.0 }& -19.6& -19.9& -19.9& -20.9& -21.0& -20.8\\
\hline
\gamename{Private Eye} & 81.9& 100.7& -23.0& 684.3& 111.9& -754.0& 0.0& \textbf { 1947.3 }\\
\hline
\gamename{Q*Bert} & \textbf { 613.5 }& 497.2& 326.3& 529.1& 565.8& 169.0& 150.0& 157.4\\
\hline
\gamename{River Raid} & 1708.9& 1438.0& 0.0& \textbf { 1904.3 }& 1309.9& 1608.6& 1070.0& 1455.5\\
\hline
\gamename{Road Runner} & 67.7& 65.2& 21.4& 42.0& 41.0& 36.2& \textbf { 900.0 }& 857.9\\
\hline
\gamename{Robotank} & 12.8& 10.1& 9.3& 10.8& \textbf { 28.7 }& 1.6& 17.0& 11.3\\
\hline
\gamename{Skiing} & -1.1& -0.7& -0.1& \textbf { -0.0 }& 0.0& 0.0& 0.0& 0.0\\
\hline
\gamename{Star Gunner} & 850.2& \textbf { 1069.5 }& 1002.2& 722.9& 769.3& 638.1& 600.0& 509.8\\
\hline
\gamename{Tennis} & -0.2& -0.1& -0.1& -0.1& -0.1& -24.0& \textbf { 0.0 }& -0.3\\
\hline
\gamename{Time Pilot} & 1728.2& 2299.5& 0.0& 2429.2& \textbf { 3741.2 }& 3458.8& 500.0& 718.7\\
\hline
\gamename{Tutankham} & 40.7& 52.6& 0.0& 85.2& \textbf { 114.3 }& 23.1& 0.0& 17.3\\
\hline
\gamename{Up and Down} & \textbf { 3532.7 }& 3351.0& 2473.4& 2475.1& 3412.6& 131.6& 550.0& 2962.9\\
\hline
\gamename{Venture} & 0.0& \textbf { 66.0 }& 0.0& 0.0& 0.0& 0.0& 0.0& 0.0\\
\hline
\gamename{Video Pinball} & 15046.8& 12574.2& 10779.5& 9813.9& 16871.3& \textbf { 20021.1 }& 705.0& 9527.9\\
\hline
\gamename{Wizard of Wor} & 1768.8& \textbf { 1981.3 }& 935.6& 945.5& 1096.2& 772.4& 300.0& 470.3\\
\hline
\gamename{Zaxxon} & 1392.0& 2069.1& 69.8& \textbf { 3365.1 }& 304.3& 0.0& 0.0& 2.0\\
\hline
\hline
Times Best & 6& 17& 1& 8& 8& 2& 9& 4\\
\hline
\end{tabular}

%% file: tables/allSearchResults.tex
\begin{tabular}{|r|r|r|r|r|}
%{|c|c|c|c|c|}
\hline
Game & Full Tree & UCT & Best Learner & Best Baseline \\ 
\hline
\hline
\gamename{Asterix} & 2135.7& \textbf { 290700.0 }& 987.3& 650.0\\
\hline
\gamename{Beam Rider} & 693.5& \textbf { 6624.6 }& 929.4& 996.0\\
\hline
\gamename{Freeway} & 0.0& 0.4& 19.1& \textbf { 22.5 }\\
\hline
\gamename{Seaquest} & 288.0& \textbf { 5132.4 }& 664.8& 451.1\\
\hline
\gamename{Space Invaders} & 112.2& \textbf { 2718.0 }& 250.1& 270.5\\
\hline
\hline
\gamename{Alien} & 784.0& \textbf { 7785.0 }& 939.2& 313.9\\
\hline
\gamename{Amidar} & 5.2& \textbf { 180.3 }& 103.4& 37.8\\
\hline
\gamename{Assault} & 413.7& \textbf { 1512.2 }& 628.0& 497.8\\
\hline
\gamename{Asteroids} & 3127.4& \textbf { 4660.6 }& 907.3& 1526.7\\
\hline
\gamename{Atlantis} & 30460.0& \textbf { 193858.0 }& 62687.0& 33058.4\\
\hline
\gamename{Bank Heist} & 21.5& \textbf { 497.8 }& 190.8& 15.0\\
\hline
\gamename{Battle Zone} & 6312.5& \textbf { 70333.3 }& 15819.7& 13000.0\\
\hline
\gamename{Berzerk} & 195.0& 553.5& 501.3& \textbf { 670.0 }\\
\hline
\gamename{Bowling} & 25.5& 25.1& \textbf { 43.9 }& 30.0\\
\hline
\gamename{Boxing} & \textbf { 100.0 }& \textbf{100.0 } & 44.0& -1.5\\
\hline
\gamename{Breakout} & 1.1& \textbf { 364.4 }& 5.2& 3.0\\
\hline
\gamename{Carnival} & 950.0& \textbf { 5132.0 }& 2323.9& 869.2\\
\hline
\gamename{Centipede} & \textbf { 125123.0 }& 110422.0& 8803.8& 16527.0\\
\hline
\gamename{Chopper Command} & 1827.3& \textbf { 34018.8 }& 1581.5& 1000.0\\
\hline
\gamename{Crazy Climber} & 37110.0& \textbf { 98172.2 }& 23410.6& 2335.4\\
\hline
\gamename{Demon Attack} & 442.6& \textbf { 28158.8 }& 520.5& 776.2\\
\hline
\gamename{Double Dunk} & -18.5& \textbf { 24.0 }& -13.1& 0.0\\
\hline
\gamename{Elevator Action} & 730.0& \textbf { 18100.0 }& 3220.6& 1040.9\\
\hline
\gamename{Enduro} & 0.6& \textbf { 286.3 }& 129.1& 25.9\\
\hline
\gamename{Fishing Derby} & -91.6& \textbf { 37.8 }& -89.5& -93.8\\
\hline
\gamename{Frostbite} & 137.2& \textbf { 270.5 }& 216.9& 175.2\\
\hline
\gamename{Gopher} & 1019.0& \textbf { 20560.0 }& 1288.3& 286.8\\
\hline
\gamename{Gravitar} & 395.0& \textbf { 2850.0 }& 387.7& 205.4\\
\hline
\gamename{H.E.R.O.} & 1323.8& \textbf { 12859.5 }& 6458.8& 712.0\\
\hline
\gamename{Ice Hockey} & -9.2& \textbf { 39.4 }& -9.5& -1.0\\
\hline
\gamename{James Bond} & 25.0& \textbf { 330.0 }& 202.8& 82.0\\
\hline
\gamename{Journey Escape} & 1327.3& \textbf { 7683.3 }& -8441.0& 0.0\\
\hline
\gamename{Kangaroo} & 90.0& \textbf { 1990.0 }& 1622.1& 498.4\\
\hline
\gamename{Krull} & 3089.2& \textbf { 5037.0 }& 3371.5& 1880.1\\
\hline
\gamename{Kung-Fu Master} & 12127.3& \textbf { 48854.5 }& 19544.0& 578.4\\
\hline
\gamename{Montezuma's Revenge} & 0.0& 0.0& \textbf { 10.7 }& 0.3\\
\hline
\gamename{Ms. Pacman} & 1708.5& \textbf { 22336.0 }& 1691.8& 505.5\\
\hline
\gamename{Name This Game} & 5699.0& \textbf { 15410.0 }& 2500.1& 3080.0\\
\hline
\gamename{Pooyan} & 909.7& \textbf { 17763.4 }& 1225.3& 540.8\\
\hline
\gamename{Pong} & -20.7& \textbf { 21.0 }& -19.0& -20.8\\
\hline
\gamename{Private Eye} & 57.9& 100.0& 684.3& \textbf { 1947.3 }\\
\hline
\gamename{Q*Bert} & 132.8& \textbf { 17343.4 }& 613.5& 169.0\\
\hline
\gamename{River Raid} & 2178.5& \textbf { 4449.0 }& 1904.3& 1608.6\\
\hline
\gamename{Road Runner} & 245.0& \textbf { 38725.0 }& 67.7& 900.0\\
\hline
\gamename{Robotank} & 1.5& \textbf { 50.4 }& 28.7& 17.0\\
\hline
\gamename{Skiing} & \textbf { 0.0 }& -0.8& 0.0& 0.0\\
\hline
\gamename{Star Gunner} & \textbf { 1345.0 }& 1207.1& 1069.5& 638.1\\
\hline
\gamename{Tennis} & -23.8& \textbf { 2.8 }& -0.1& 0.0\\
\hline
\gamename{Time Pilot} & 4063.6& \textbf { 63854.5 }& 3741.2& 3458.8\\
\hline
\gamename{Tutankham} & 64.1& \textbf { 225.5 }& 114.3& 23.1\\
\hline
\gamename{Up and Down} & 746.0& \textbf { 74473.6 }& 3532.7& 2962.9\\
\hline
\gamename{Venture} & 0.0& 0.0& \textbf { 66.0 }& 0.0\\
\hline
\gamename{Video Pinball} & 55567.3& \textbf { 254748.0 }& 16871.3& 20021.1\\
\hline
\gamename{Wizard of Wor} & 3309.1& \textbf { 105500.0 }& 1981.3& 772.4\\
\hline
\gamename{Zaxxon} & 0.0& \textbf { 22610.0 }& 3365.1& 2.0\\
\hline
\hline
Times Best & 4& 45& 3& 3\\
\hline
\end{tabular}